# From dots to faces: Individual differences in visual imagery capacity predict the content of Ganzflicker-induced hallucinations

Ana Chkhaidze[1], Reshanne R. Reeder[2], Connor Gag[3], Anastasia Kiyonaga[1], Seana Coulson[1,4]

[1]Department of Cognitive Science, University of California, San Diego (USA)
[2]Department of Psychology, Institute of Population Health, University of Liverpool (UK)
[3]Department of Computer Science, University of California, San Diego (USA)
[4]Kavli Institute for Brain and Mind, San Diego (USA)

Correspondence to: Seana Coulson (scoulson@ucsd.edu)

## Abstract

A rapidly alternating red and black display known as Ganzflicker induces visual hallucinations that reflect the generative capacity of the visual system. Individuals vary in their degree of visual imagery, ranging from absent to vivid imagery. Recent proposals suggest that differences in the visual system along this imagery spectrum should also influence the complexity of other internally generated visual experiences. Here, we used tools from natural language processing to analyze free-text descriptions of hallucinations from over 4,000 participants, asking whether people with different imagery phenotypes see different things in their mind's eye during Ganzflicker-induced hallucinations. Topic modeling of descriptions revealed that strong imagers described complex, naturalistic content, while weak imagers reported simple geometric patterns. Using crowd-sourced sensorimotor norms, we also found that participants with stronger imagery used language with richer perceptual associations. These findings may reflect individual variation in coordination between early visual areas and higher-order regions relevant for the imagery spectrum.

## From dots to faces: Individual differences in visual imagery capacity predict the content of Ganzflicker-induced hallucinations

In 1819, the physiologist Jan Purkinje made a striking observation: While waving his hand between his closed eyes and sunlight, he perceived "beautiful regular figures that are initially difficult to define but slowly become clearer" (Purkinje, 1819). His drawings of these experiences—geometric lattices, radiating patterns, and spiraling forms—represent the first scientific documentation of what we now call *flicker-induced visual hallucinations*: visual percepts that arise not from external stimuli but from the brain's response to rhythmic light stimulation without pharmacological intervention (Hewitt et al., 2025). Flicker stimulation has become a widely used experimental method for non-invasively and non-pharmacologically inducing *altered states of consciousness* (ASCs): transient shifts in subjective experience that differ qualitatively from ordinary waking awareness (Fort et al., 2025). From Purkinje's early introspections to modern laboratory studies (Fig. 1), the experienced visuals in flicker paradigms have been so consistent that they have been codified as "form constants," a universal vocabulary for simple visual hallucinations (Rule et al., 2011). This consistency suggests that rather than transient perceptual noise, flicker-induced hallucinations reflect the intrinsic architecture of the human visual system (Bressloff et al., 2001; 2002). As such, they offer a unique experimental window into other forms of internally generated visual experiences, including visual mental imagery, a fundamental yet poorly understood cognitive trait.

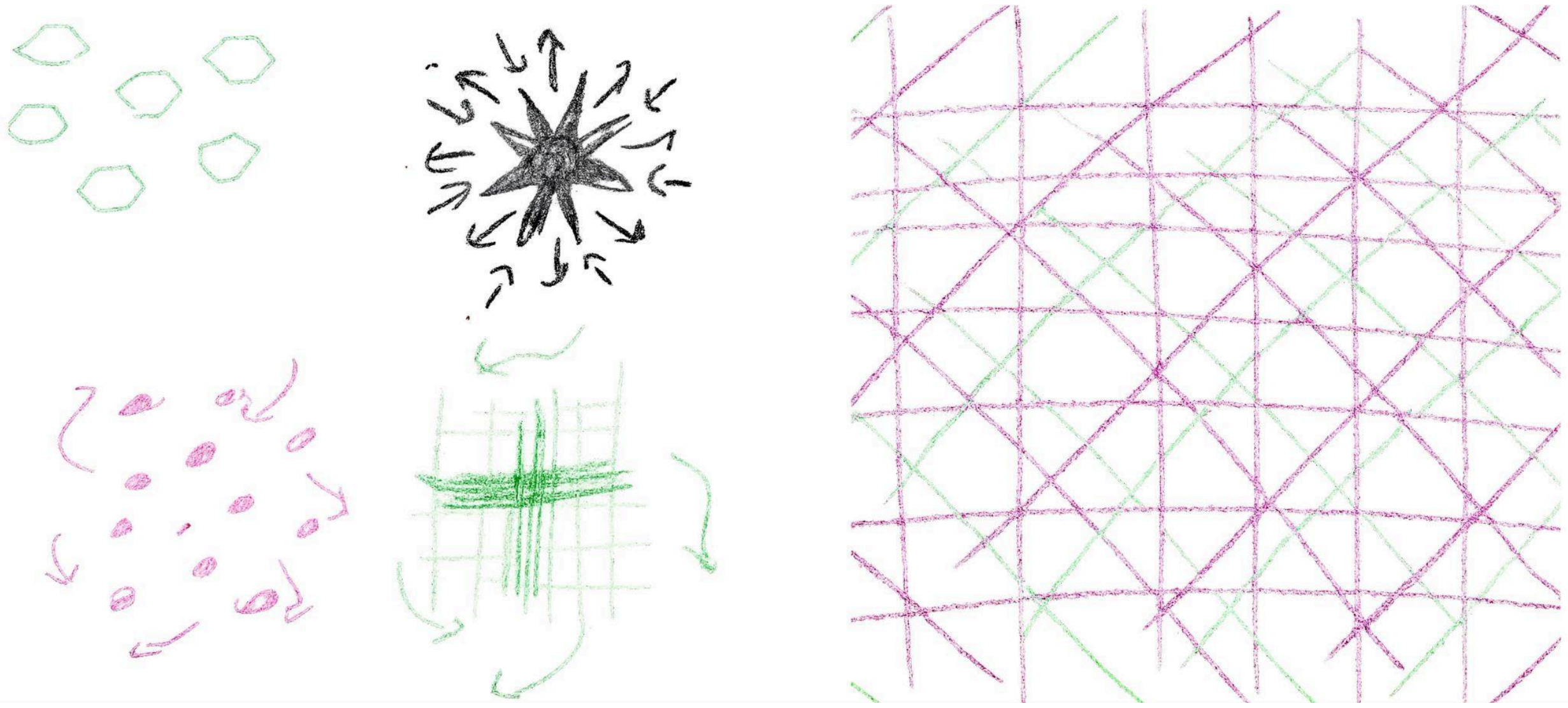

**Figure 1. Ganzflicker-induced hallucination drawings from a pilot participant in our lab.** Verbal report: *"Initially sort of a grid-like pattern, but with certain parts being more emphasized than others, and moving around across the screen. Then at some point sort of an array of pulsating dots, also moving around the screen. Then something like starbursts. Occasionally the grid-like pattern would return, and during some periods there was just a sort of non-descript dynamic patterns moving around quickly. At one point more toward the end, I was seeing something similar to chemical molecular structures. All of these were in constant motion and fluctuation."*

Mental imagery, the ability to internally simulate sensory experiences, is one of the most familiar phenomenological aspects of the human experience. Argued to play a central role in core cognitive and affective functions (Pearson, 2019), visual mental imagery has been shown to relate to working memory (Keogh & Pearson, 2014), mind wandering (Christian et al., 2013), navigation (Bird et al., 2012), and language comprehension (Muraki et al., 2023). However, visual mental imagery capacity varies dramatically across individuals along what we term the *visual imagery spectrum* (referred to below simply as the imagery spectrum), ranging from aphantasia (complete absence of visual imagery) to hyperphantasia (strong, lifelike mental pictures). Such diversity raises fundamental questions about cognitive universality: if people differ in how they internally represent sensory information, then the underlying mechanisms of the cognitive phenomena that draw on these representations may also differ systematically. Indeed, recent work demonstrates that imagery differences can predict memory recall (Bainbridge et al., 2021), emotional reactivity (Wicken et al., 2021), and working memory strategies (Reeder et al., 2024), suggesting that understanding individual differences across the imagery spectrum may be crucial for developing more inclusive theories of human cognition (Lupyan et al., 2023). While research has extensively documented that imagery vividness varies across individuals (Zeman, 2024), we know remarkably little about what people with different imagery phenotypes 'see'—that is, the content and structure of these internally generated visual experiences. Here we used tools from natural language processing to analyze participants' descriptions of their visual hallucinations, asking whether people across the imagery spectrum see different things in their mind's eye.

**The layered model of visual imagery**

Historically, imagery research has predominantly focused on scalar measures of vividness (e.g. the Vividness of Visual Imagery Questionnaire; Marks, 1973) rather than the compositional elements that constitute mental imagery (e.g., imagery manipulation and precision; Kosslyn et al., 1984). Initial neuroimaging studies of the imagery spectrum focused on early visual cortex, revealing that imagery vividness and precision correlate with the pattern and extent of activation in V1 during imagery tasks (Albers et al., 2013; Bergmann et al., 2016; Cui et al., 2007; Lee et al., 2012). However, more recent work has extended this investigation to the entire brain, confirming differences in early visual cortex when decoding imagined content (Cabbai et al., 2024), and revealing activation differences in insula (Cabbai et al., 2024; Silvanto & Nagai, 2025), frontoparietal control networks (Spagna et al., 2021), and the fusiform gyrus (Liu et al., 2025), reflecting the distributed, multi-level nature of mental imagery processes.

Importantly, this distributed activity may translate into individual variation not just in imagery vividness, but in the content and structural complexity of what people can internally represent. While individuals across the imagery spectrum may similarly represent low-level visual features (lines, edges, spatial patterns), those with weaker imagery or aphantasia may have reduced capacity for integrating these elements into complex, semantically meaningful images due to differences in top-down connectivity between control networks and visual areas during imagery (Milton et al., 2021; Liu et al., 2025). For example, when imagining a butterfly landing on a flower, early visual cortex might encode basic features like curved edges, color patches, and black lines across all individuals, but the successful construction of a vivid butterfly image may require higher-order semantic and object processing areas to integrate these features into meaningful shapes, along with the recruitment of color-selective areas to contribute the rich hues in the scene, as well as coordination via frontoparietal control networks to bind these elements into a unified percept (Dijkstra et al., 2019; Mechelli et al., 2004; Spagna et al., 2021). This coordination may be compromised in weaker imagers, resulting in fragmented or impoverished butterfly imagery despite intact basic feature representations. We refer to this framework as the *layered model of visual imagery* in which individuals across the imagery spectrum similarly represent low-level visual features in the early visual cortex, but differ in their capacity to integrate these elements into complex, semantically meaningful representations (for details see Fig. 2).

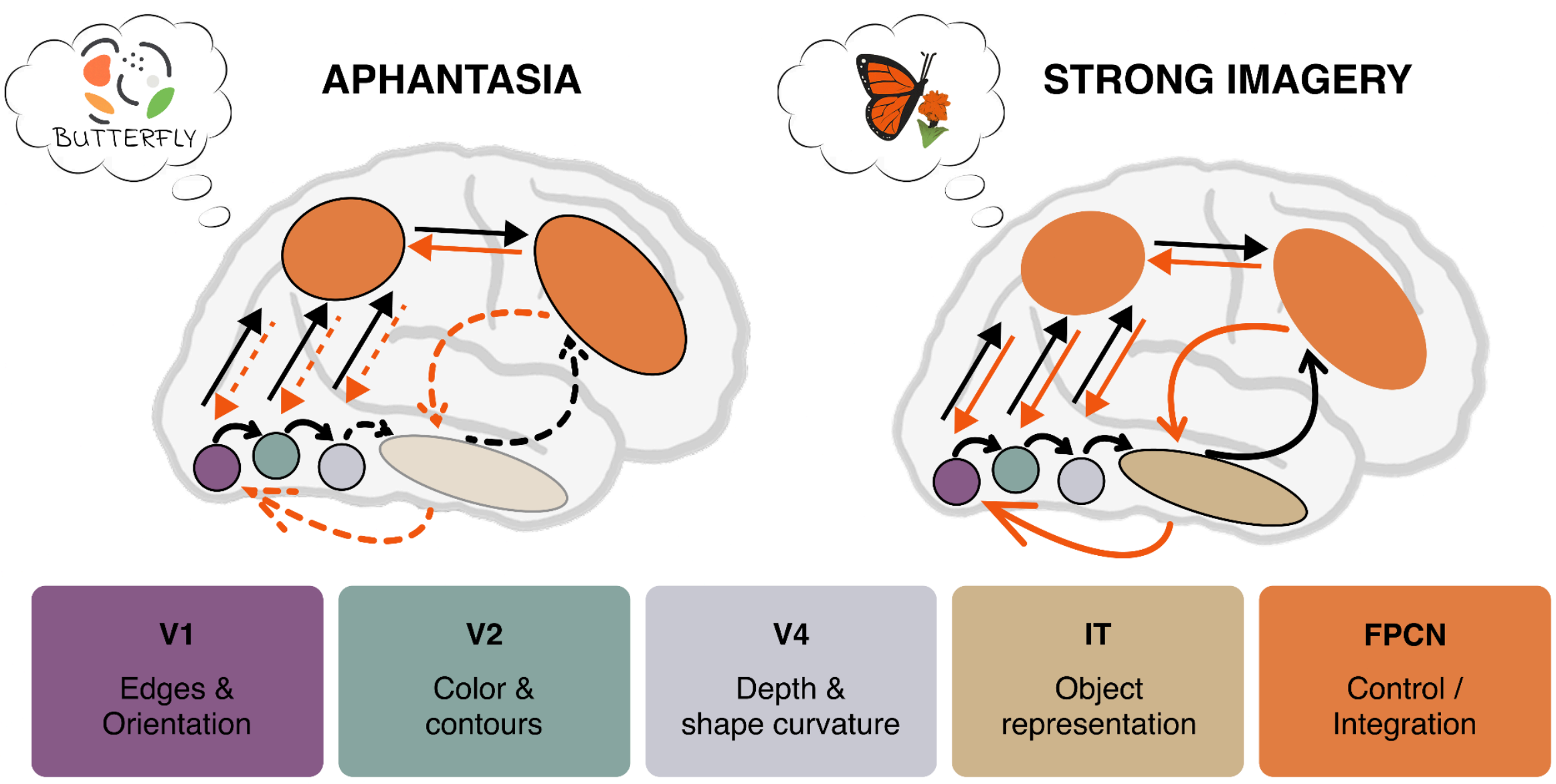

**Figure 2. Layered model of visual imagery.** Proposed differences in connectivity across the visual hierarchy during internally generated visual experiences between aphantasia and strong imagery. Black arrows indicate bottom-up feedforward connections; orange arrows indicate top-down modulatory connections. Solid lines represent strong connectivity; dashed lines represent reduced connectivity. Both individuals with aphantasia and with moderate to strong imagery show intact bottom-up flow from early visual cortex (EVC: V1, V2, V4) through parietal to frontal regions, explaining why people with no self-reported imagery still experience simple visuals during Ganzflicker-induced hallucinations (Königsmark et al., 2021; Reeder, 2022). The key difference lies in the efficiency of coordinated activity across the hierarchy. In strong imagers, a combination of strong bidirectional connectivity between prefrontal areas and inferotemporal (IT) cortex and robust top-down modulation of visual cortical activity via the fronto-parietal control network (FPCN) enable integration of low-level features into coherent, semantically meaningful imagery. In aphantasia, these connections are weaker, limiting conscious imagery to simple content.

**Individual differences in imagery and the induction of visual hallucinations**

If individual differences in visual imagery reflect variation in the hierarchical coordination of visual representations, then these differences should manifest not only during voluntary imagery but also in other situations where internally generated content shapes perception, such as during flicker-induced hallucinations (Reeder et al., 2024). Since Purkinje's pioneering forays, experimental methods to induce visual hallucinations have evolved, systematized, and diversified. Visual hallucinations can be successfully induced through both sensory deprivation (e.g., blindfolding; Merabet et al., 2004) and sensory stimulation (e.g., stroboscopic light stimulation; Hewitt et al., 2025). A combination of both, the *Ganzfeld* paradigm involves presenting participants with a uniform colored field, inducing hallucinations as the brain generates structure in the absence of patterned input (Metzger, 1929; Pütz et al., 2006; Schmidt & Prein, 2019; Schmidt et al., 2020; Wackermann et al., 2008). Stroboscopic light stimulation by contrast, delivers bright rhythmic stimulation, typically white light to closed eyes, producing hallucinations possibly through neural entrainment in the visual cortex (as reviewed in Hewitt et al., 2025). The *Ganzflicker*—a special case of the flicker-based paradigm—combines elements of both approaches: like stroboscopic light stimulation, it uses rhythmic flicker to drive visual experience; like Ganzfeld, it employs uniform chromatic stimulation, though unlike previous suggestions, this does not have to fill the entire visual field to elicit hallucinations (Reeder, 2022). Specifically, the term refers to open-eye viewing of two uniform chromatic fields (typically red and black) alternating rhythmically (Königsmark et al., 2021). All these induction methods can successfully lead to ASCs (Fort et al., 2025), though the phenomenological character of the resulting experiences may vary (Fort et al., 2025; Shenyan et al., 2024).

Simple hallucinations—geometric shapes, color alterations, and elementary shapes—are the most commonly reported experiences in these light-induced hallucinations, across both Ganzfeld and flicker-based paradigms (Allefeld et al., 2011; Bartossek et al., 2021; Hewitt et al., 2025; Pistolas & Wagemans, 2025; Pütz et al., 2006; Rule et al., 2011; Schmidt & Prein, 2019; Schmidt et al., 2020; Wackermann et al., 2008). These experiences have been suggested to reflect the intrinsic dynamics of early visual cortex as computational models demonstrate that conditions of either rhythmic stimulation or reduced sensory input result in these patterned representations due to lateral connectivity and retinotopic organization in V1 (Bressloff et al., 2001, 2002; Ermentrout & Cowan, 1979).

Despite the predominance of simple hallucinations, complex hallucinatory content has also been reported in both Ganzfeld (Pütz et al., 2006; Shenyan et al., 2024) and flicker-induced paradigms (Allefeld et al., 2011; Königsmark et al., 2021; Reeder, 2022; Shenyan et al., 2024; Schwartzman et al., 2019) (Fig. 3). Complex hallucinations involve recognizable semantic content, like faces, human figures, animals, objects, landscapes, and structured scenes. Notably, the likelihood of experiencing complex versus simple hallucinations appears to depend in part on the induction method. Shenyan et al. (2024) found that complex hallucinations were relatively more likely during Ganzfeld than Ganzflicker, suggesting that bottom-up rhythmic stimulation of the early visual cortex may not be sufficient to generate complex content. Instead, complex hallucinations may require additional involvement of higher-level visual areas—either through feedforward propagation up the visual hierarchy or through top-down interpretative processes (see e.g., Shenyan et al., 2024). Consistent with a role for top-down processing, Pistolas and Wagemans (2025) documented complex imagery (fish, sharks, ocean scenes) that appeared to be shaped by contextual cues such as illumination color and ambient sound. However, only a portion (~20%) of participants were susceptible to such context-driven influences. Complex hallucinations also occur in flicker paradigms: some participants undergoing stroboscopic light stimulation have reported semantically rich scenes such as wheat fields, faces of loved ones, and birds flying into space (Beauté et al., 2025). Thus, hallucination complexity appears to depend on more than just the induction method and may also vary as a function of stable individual traits such as visual imagery capacity (Beauté et al., 2025).

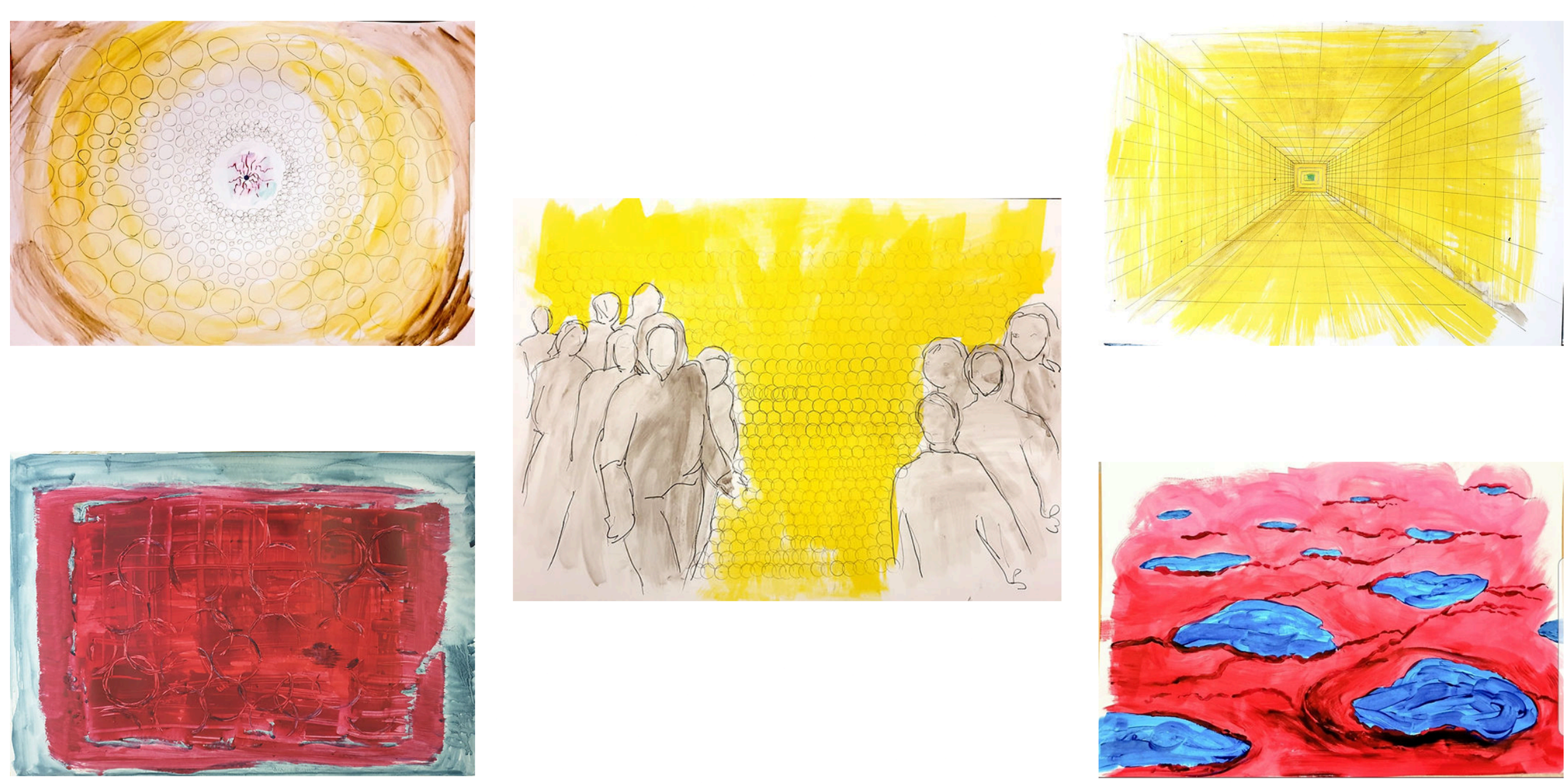

**Figure 3. Ganzflicker-induced simple and complex hallucinations.** Drawings by an artist based on her firsthand Ganzflicker experiences, illustrating both simple and complex visual hallucinations, including colored geometric patterns, a tunnel, and human figures. Reproduced with permission; drawings © Kirsten Baron.

Indeed, in recent years visual imagery capacity has emerged as a reliable predictor of flicker-induced hallucinations. Large-scale studies using the Ganzflicker paradigm show that individuals with stronger imagery report more frequent and more vivid hallucinations during Ganzflicker stimulation (Königsmark et al., 2021; Reeder, 2022). If complex hallucinations depend on layered mechanisms similar to those supporting voluntary visual imagery, then individuals with stronger imagery capacity should also be more likely to experience complex content. When asked to categorize their hallucinations using fixed response options, participants with strong imagery were more likely to report complex hallucinations than those with weak imagery (Reeder, 2022).

While these studies established the first empirical link between visual imagery capacity and Ganzflicker-induced hallucinations, closed-ended response formats cannot fully reveal the experiential content of hallucinations. The layered model of imagery offers a testable prediction about hallucination content across the imagery spectrum: individuals with stronger imagery should report more complex, naturalistic hallucination content, while those with weaker imagery should predominantly report the simple geometric patterns linked to activity in the early visual cortex. In other words, the same processing differences that distinguish imagery phenotypes should be visible in the content of what people 'see' during Ganzflicker—a prediction that can be tested by analyzing the semantic structure of open-ended phenomenological reports.

**The present study**

To characterize variability in hallucination content across the imagery spectrum, we conducted a secondary analysis of an open dataset containing over 6,000 participants' free-text descriptions of their experiences following Ganzflicker stimulation (Reeder, 2022). After the Ganzflicker exposure, participants provided self-reported visual imagery vividness ratings to determine their position on the imagery spectrum, and answered structured questions about their experiences during Ganzflicker (Fig. 4a). While Reeder (2022) focused primarily on these structured questionnaire data, participants also provided rich narrative reports describing their visual hallucinations (see Supplementary Material 1 for example descriptions), a corpus that has not previously been analyzed for content differences across the imagery spectrum. Analyzing the content and structure of these free-text descriptions requires a methodological approach that can accommodate the variability and richness of open-ended phenomenological reports while enabling systematic, replicable analysis. To address this challenge, we employed two complementary methods (Fig. 4c).

First, we aimed to uncover the phenomenological content of Ganzflicker-induced hallucinations, asking whether it changes across the imagery spectrum. To locate latent themes in participants' descriptions, we used MOSAIC (Mapping of Subjective Accounts into Interpreted Clusters), a topic modeling pipeline that relies on machine learning to identify coherent themes in text corpora, allowing data-driven discovery of hallucination content (Beauté et al., 2025). Following prior literature (Hewitt et al., 2025; Reeder, 2022), we expected a range of simple and complex themes to emerge. We hypothesized that participants with stronger imagery would describe more complex, naturalistic content (faces, scenes, structured hallucinations), while those with weaker imagery would report simpler, geometric patterns, arising from the individual differences in brain organization posited in the layered model of imagery.

Second, we used the Lancaster Sensorimotor Norms (Lynott et al., 2020) to assess explicit sensorimotor language in participants' descriptions. These norms provide crowdsourced ratings of how strongly words evoke different sensory and motor experiences—for example, rating "thunder" as highly auditory, "velvet" as highly tactile, "kick" as foot-related, or "clap" as hand-related. While topic modeling may reveal thematic content at the level of whole descriptions, the Lancaster norms complement this analysis by indexing explicit sensorimotor vocabulary at the word level. Moreover, as they are derived from human ratings, they offer a psychologically grounded measure with established links to behavioral indices of word processing, including word naming and the lexical decision task (Connell & Lynott, 2012), as well as neural responses to words in the property verification task (Vinaya et al., 2025). Crucially, these norms allowed us to identify which sensory and motor dimensions emerge in Ganzflicker-induced hallucinations and test whether they vary systematically across the imagery spectrum. We expected participants with higher imagery vividness scores to use language with stronger sensorimotor associations, reflecting richer internal simulation.

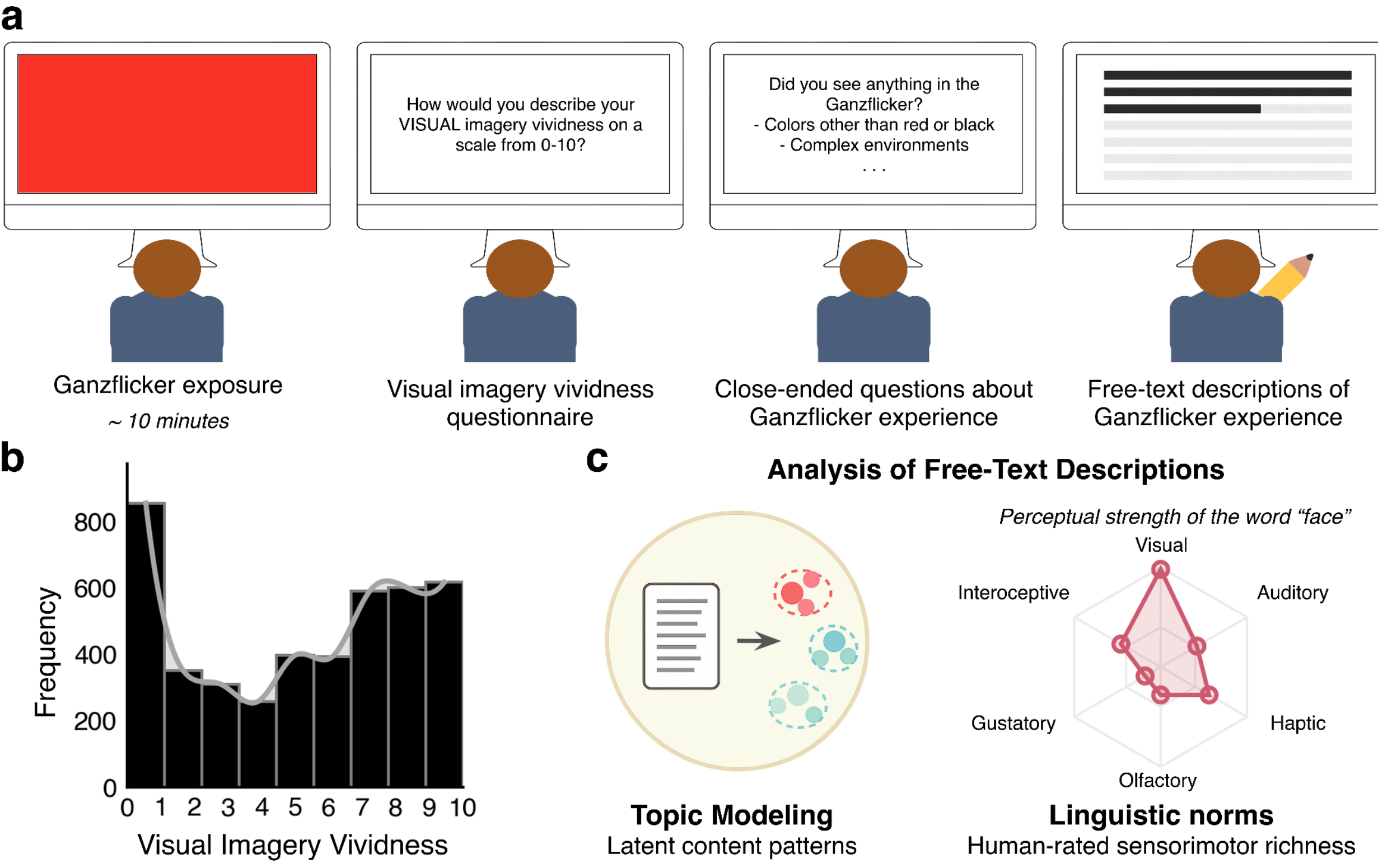

**Figure 4. Ganzflicker task and analytical approach. a,** Experimental procedure from Reeder (2022). Participants viewed red-black flicker for ~10 minutes, completed a demographics questionnaire and rated their visual imagery vividness (0-10); afterwards, they answered whether they saw anything during Ganzflicker and provided free-text descriptions of what they saw, followed by additional closed-ended questions about their experience. **b,** Distribution of visual imagery vividness ratings across participants in this study. Bars show participant counts at each discrete vividness rating; the curve shows the overall shape of the distribution. **c,** Analysis pipeline: topic modeling to identify semantic themes and human-rated linguistic norms to quantify sensorimotor content in hallucinations.

## Materials and methods

### Participants

We used a dataset from a Ganzflicker study (Reeder, 2022) with 6,664 individuals who underwent 10 minutes of continuous Ganzflicker stimulation and reported their experiences, alongside visual imagery vividness scores. Participants were recruited via a popular science article published in The Conversation describing the Ganzflicker phenomenon (Reeder, 2021), which was republished by at least 14 media outlets including Science Alert, The Daily Beast, and Big Think, all of which contained a link to the Ganzflicker study hosted on Google Forms. This widespread media coverage led to a large, international volunteer sample, with the majority of readers based in the USA (73%), followed by the UK (5.2%), Canada (4.1%), and Australia (3.3%). 68.1% of participants reported their gender as male, 27.6% as female, 2.4% as other, and 1.9% preferred not to say. 5,707 individuals reported their age (M = 38.07, SD = 15.08). The sample also included a higher-than-typical proportion of individuals at the extremes of the imagery spectrum, including self-reported aphantasia (Fig. 4b). We excluded participants whose descriptions were missing or who provided written descriptions in a language other than English, leaving 4,365 participants. As reported by Reeder (2022), the study was approved by the ethics committee of the medical faculty of Otto-von-Guericke University and all participants gave informed consent.

### Procedure

Participants viewed a Ganzflicker stimulus that alternated between red and black at 7.5 Hz for approximately 10 minutes. They were instructed to watch Ganzflicker on full-screen and under dim lighting, accompanied by white noise to homogenize the acoustic environment, and to avoid distractions. Approximately 73% of participants reported compliance with the white noise recommendation; however, the proportion of complex versus simple hallucinations did not differ between those who used white noise and those who did not (Königsmark et al., 2021; Reeder, 2022), suggesting that auditory masking did not substantially influence hallucination content. Further details on environment control instructions are reported in Reeder (2022).

After the stimulus presentation, participants completed a demographic survey. To situate them on the imagery spectrum, they rated their visual imagery vividness on a single item ("How would you describe your VISUAL imagery vividness on a scale from 0 (no mental imagery) to 10 (as vivid as real perception)). Participants also answered closed-ended questions assessing their Ganzflicker experiences, and an open-ended free-text description of what they saw during exposure. In this paper, we focus on these free-text descriptions.

### Topic Modeling

To uncover latent themes in participants' descriptions of Ganzflicker-induced hallucinations, we implemented a multi-step topic modeling pipeline based on BERTopic (Grootendorst, 2022), an unsupervised method that integrates transformer-based embeddings, dimensionality reduction, density-based clustering, and class-based term weighting (Fig. 5). After clustering, each topic's most informative terms were extracted using class-based Term Frequency-Inverse Document Frequency (c-TF-IDF), which identifies words that best characterize a given cluster relative to the rest of the corpus. These keyword sets were used as the primary basis for human-interpretable topic labels. To further reduce researcher expectancy bias, we also employed GPT-4o-mini to generate independent topic labels based on the same representative keywords and exemplar sentences. Human- and model-generated labels showed close alignment (see Table 1). A more explicit description of the pipeline is detailed in Fig. 5.

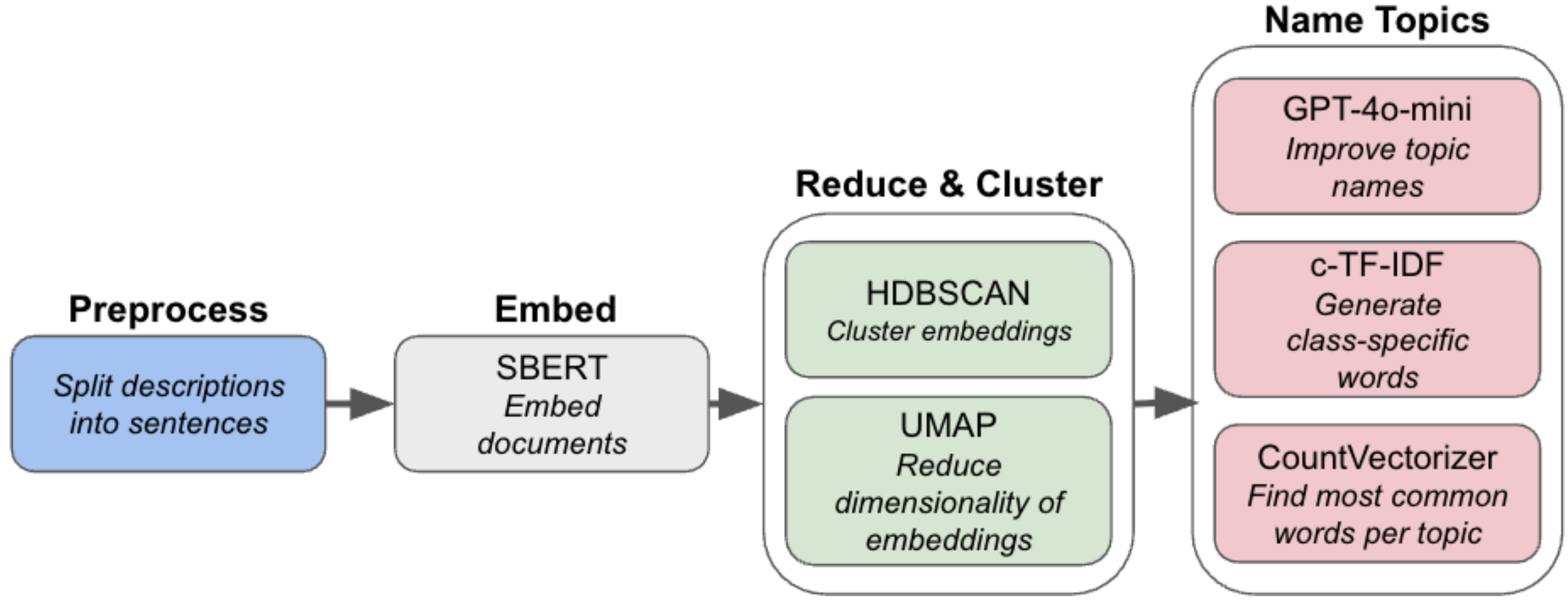

**Figure 5. Topic modeling pipeline.** Topic modeling pipeline: Free-text hallucination descriptions were split into sentences, embedded using Sentence-BERT, reduced with UMAP, and clustered using HDBSCAN. Topics were labeled using a combination of most-frequent terms, c-TF-IDF, and GPT-4o-mini summarization.

**Sentence-level tokenization and preprocessing.** Each hallucination report was first split into sentences using sent_tokenize() from the Natural Language Toolkit (NLTK) (Bird et al., 2009), which applies the *Punkt* algorithm (Kiss & Strunk, 2006)—an unsupervised model that uses punctuation and contextual cues to identify sentence boundaries. This step allowed us to capture distinct experiential elements that often co-occur within a single report. Treating entire descriptions as a single document would have conflated diverse perceptual content under one topic. The resulting sentence-level documents were lightly cleaned—converted to lowercase and stripped of excess whitespace—but no stemming, lemmatization, or stopword removal was performed. This minimal preprocessing ensured the preservation of contextual information important for semantic modeling, resulting in a corpus of over 10,000 sentence-level documents.

**Sentence embeddings and dimensionality reduction**. We generated semantic embeddings using Sentence-BERT (Reimers & Gurevych, 2019), which produces 384-dimensional vector representations optimized for capturing semantic similarity in short texts. This embedding process preserves subtle differences in meaning and context, allowing the model to distinguish between similar phrases with different phenomenological implications (e.g., “I felt time slow down” vs. “I lost track of time”).

To prepare the embeddings for clustering, we applied Uniform Manifold Approximation and Projection (UMAP) to reduce their dimensionality (McInnes et al., 2018). UMAP preserves both local and global structure in the data, helping maintain the semantic relationships between sentences in the lower-dimensional space. We configured the reduced dimensionality (n_components = 10), the neighborhood size used to balance local versus global structure (n_neighbors = 15), and the minimum distance between embedded points (min_dist = 0.1) so as to yield an embedding structure that supported coherent and interpretable topic clustering.

**Clustering with HDBSCAN.** The reduced sentence embeddings were clustered using Hierarchical Density-Based Spatial Clustering of Applications with Noise (HDBSCAN) (McInnes et al., 2017; Campello et al., 2013), a non-parametric algorithm that identifies clusters of varying density without requiring a predefined number of clusters. This feature is particularly advantageous when analyzing naturalistic text, where the number and size of coherent experiential categories are unknown. We set the minimum cluster size to 30, allowing the model to discover a diverse but stable set of themes while treating small or noisy clusters as outliers. Sentences not confidently assigned to any topic were labeled as outliers (topic = –1) and excluded from topic-based analyses to ensure the coherence and interpretability of topic clusters. A total of 32 sentences (<0.3% of the corpus) were labeled as outliers; these were short Turkish-language descriptions that had not been flagged during automated language detection and were excluded from topic analyses. The final model identified 28 topics spanning a wide range of visual and conceptual content, ranging from simple geometric shapes to complex scenery (Fig. 6, Table 1).

**Topic extraction and coherence evaluation**. To identify the most distinctive terms characterizing each topic, we applied c-TF-IDF (Grootendorst, 2022). This weighting scheme highlights words that are frequent within a topic and infrequent across other topics. Unlike standard TF-IDF, which considers individual documents, c-TF-IDF treats all documents within a topic as a single pooled entity, allowing for robust extraction of topic-defining terms. Specifically, our pipeline employed the BM25+ variant of c-TF-IDF, which provides additional smoothing and has been shown to improve performance on sparse or short-text data.

To evaluate the interpretability of topics, we computed the $C_v$ coherence score (Röder et al., 2015), a widely used metric that assesses the degree of semantic similarity among the top-ranked words within each topic. $C_v$ combines normalized pointwise mutual information (NPMI), cosine similarity, and a sliding window approach to assess word co-occurrence. Our model achieved a coherence score of 0.48, which falls within the commonly reported range of 0.4 to 0.7 in topic modeling studies involving short, noisy, or experiential texts (Röder et al., 2015; O’Callaghan et al., 2015). Topic structure was further examined using a 2D topic embedding map (Fig. 6) and a hierarchical dendrogram (Fig. S1), which visually confirmed the separability and cohesion of extracted themes.

**Topic labeling**. To reduce researcher bias—i.e., the tendency for manual labeling to be influenced by theoretical priors—and to enhance interpretability of the keywords, we used *GPT-4o-mini* (OpenAI, 2024) for LLM-assisted topic labeling. As outlined by Beauté et al. (2025), this method offers a scalable and unbiased alternative to manual labeling for large phenomenological corpora. For each topic, we supplied the model with its top 10–15 c-TF-IDF keywords and a few representative sentences, prompting it to generate concise, human-readable topic labels. We also independently reviewed the keywords and representative example sentences for each topic and manually assigned labels. The two approaches produced highly consistent results, differing mainly in surface phrasing (e.g., "city skyline" vs. "cityscape views"). Topic modeling figures (Figs. 5, 6, S1, S2, & S3) reflect the manually-generated labels; Table 1 includes both human-generated and LLM-generated labels for comparison.

**Participant-level topic features**. Each sentence was assigned both a hard topic label and a soft probability distribution over all topics. We used BERTopic's standard hard-label assignment, where the most probable topic for each sentence was determined as the argmax of the topic-probability vector $p$(topic | sentence). In addition, BERTopic provides the full topic-probability distribution for each sentence, indicating the likelihood that it belongs to each topic. To construct participant-level features, we aggregated these probabilities by identifying, for each participant × topic pair, the maximum probability that the model assigned to that topic across all of the participant's sentences. This maximum-based aggregation preserves sensitivity to distinct, high-confidence topic expressions and prevents unrelated sentences from diluting the signal. The resulting participant-by-topic matrix contained continuous topic salience scores, where each cell reflected the strongest evidence that a given participant expressed that experiential theme (scores near 1 indicating strong expression, and scores near 0 indicating absence). These salience values indexed the presence and prominence of each topic at the participant level and served as inputs for subsequent regression and classification models testing whether topic structure predicted self-reported imagery vividness.

### Predicting imagery vividness from hallucination content

To test whether the derived topic structure could distinguish individuals with different imagery phenotypes, we modeled the relationship between hallucination topics and participants' vividness scores. We used Lasso regression with L1 regularization (Tibshirani, 1996) to predict participants' self-reported vividness scores based on their hallucination content. Lasso regression was selected here for its ability to identify sparse, stable predictions in high-dimensional and potentially collinear feature spaces. Our analytic goal was not prediction per se, but interpretability: to identify robust topic-vividness associations that reflect the content structure of visual hallucinations across the imagery spectrum.

The predictor variables were the 28 z-scored topic probabilities obtained from the topic modeling analysis. The dataset was randomly split into 80% training and 20% testing subsets. The optimal regularization parameter ($\alpha$) was selected via 10-fold cross-validation, testing 100 logarithmically spaced values between 0.001 and 10. All analyses were conducted in *Python* (Python Software Foundation, 2023) using scikit-learn (Pedregosa et al., 2011).

To further characterize hallucination content across distinct imagery phenotypes, we categorized participants into three imagery groups based on vividness scores: weak (0–3), moderate (4–7), and strong (8–10), following conventions in prior work (Reeder, 2022; Zeman et al., 2020). These bins approximate the distinction between aphantasia, typical imagery, and hyperphantasia, respectively. Final group sizes were: weak = 1,515, moderate = 1,634, and strong = 1,216. This approach allowed us to investigate which topics constituted the "fabric" of hallucination content in each imagery group and visualize their distinctive content patterns.

We trained three separate Lasso-regularized binary classifiers to predict imagery group membership from hallucination content, represented as standardized topic probabilities. Each classifier was trained to distinguish one imagery group (weak, moderate, or strong) from all others using stratified train-test splits (80% training, 20% testing). We used LogisticRegressionCV class from scikit-learn (Pedregosa et al., 2011) with the following parameters: penalty='l1', solver='saga', scoring='f1', cv=10, max_iter=10,000, and class_weight='balanced'. The regularization parameter *C* was optimized through 10-fold cross-validation across 30 logarithmically-spaced values between 0.01 and 100. Class balancing was implemented during training by assigning higher weights to under-represented classes to ensure model sensitivity to signals in minority groups even when overall accuracy remained limited.

Non-zero coefficients from Lasso-regularized classifiers represent the most robust content–vividness relationships that survived both regularization penalties and class imbalance challenges. To assess coefficient stability, we performed 1,000 bootstrapped model iterations for each imagery group using the *resample* function from scikit-learn (Pedregosa et al., 2011). In each iteration, we resampled the standardized feature matrix and target labels, fitted a logistic regression model with L1 penalty using the previously determined optimal regularization parameter, and recorded which features received non-zero coefficients. We retained only topics that survived regularization in at least 60% of iterations as our threshold for interpretability.

To assess model performance, we used the F1 score—the harmonic mean of precision and recall. F1 is a widely used evaluation metric in imbalanced classification settings where sensitivity to the minority class is important. This metric is particularly well-suited to contexts where the goal is not maximizing raw accuracy, but rather understanding which features meaningfully distinguish categories while balancing false positives and false negatives (Saito & Rehmsmeier, 2015). To evaluate the statistical significance of model performance, we conducted permutation tests (Nichols & Holmes, 2002) for each classifier. In these tests, we randomly shuffled the imagery group labels—thereby breaking any systematic relationship between hallucination content and vividness category—and retrained the model 1,000 times on each shuffled dataset. This procedure generated a null distribution of F1 scores under the assumption that no true association exists. We then compared the F1 score from the original model to this null distribution to evaluate whether the model's performance reflected real structure in the data.

**Sensorimotor content analysis**

To supplement our computational study with human judgment perspective, we conducted a complementary study examining the perceptual and embodied content of hallucination descriptions using the Lancaster Sensorimotor (LS) norms (Lynott et al., 2020) as described in Chkhaidze et al. (2023) (see Fig. 8a for the pipeline).

**Description preprocessing**. We first preprocessed hallucination descriptions using standard natural language processing techniques in *Python* with NLTK (Bird et al., 2009). This involved: (1) tokenization, (2) removal of punctuation and conversion to lowercase, (3) removal of English stopwords using the NLTK stopwords corpus, (4) word lemmatization to reduce words to their base forms, and (5) spell correction using the pyspellchecker library (Barrus, 2021). Description length was calculated as the total number of tokens after preprocessing.

**Quantifying descriptions using LS Norms**. The LS Norms provide human-derived ratings for around 40,000 English words across 11 dimensions: six perceptual modalities (visual, auditory, gustatory, olfactory, haptic, and interoceptive) and five motor/action dimensions (foot, hand, head, mouth, and torso). For each word, human raters indicated how strongly they experience that concept through each sensory modality or through actions involving each body part. In addition to these individual modality ratings, the LS Norms include two composite scores: *perceptual strength* and *action strength*. These reflect the maximum strength value assigned to the dominant modality for each word—e.g., if a word is rated highest on the *visual* dimension, its perceptual strength is that visual rating. This approach is designed to capture the most salient sensory or motor modality through which a concept is typically experienced, and is theorized to reflect conceptual concreteness (Banks & Connell, 2022).

For each hallucination description, we matched preprocessed words to the LS database and calculated average scores across all six perceptual, five motor LS dimensions, and two composite measures. Descriptions with fewer than three valid LS norm-matched words (~7%) were excluded and we ended up with 4,057 participants for this analysis.

**Regression models**. We asked three research questions. First, we tested whether the overall perceptual and action richness of hallucination descriptions predicted participants' self-reported imagery vividness. To do this, we fit a generalized linear model (GLM) predicting vividness from *perceptual strength* and *action strength*. The model was formulated as:

$$\text{visual vividness}_i = \beta_0 + \beta_1 \cdot \text{perceptual strength}_i + \beta_2 \cdot \text{action strength}_i + \epsilon_i$$

Second, to clarify which specific perceptual dimensions were driving the overall perceptual strength effect, we fit a GLM including all six LS perceptual modality scores as simultaneous predictors:

$$\text{visual vividness}_i = \beta_0 + \beta_1 \cdot \text{visual}_i + \beta_2 \cdot \text{auditory}_i + \beta_3 \cdot \text{gustatory}_i + \beta_4 \cdot \text{olfactory}_i + \beta_5 \cdot \text{interoceptive}_i + \epsilon_i$$

Lastly, we asked whether the overall action strength effect was similarly driven by any specific motor modality. We fit a separate GLM using the five motor dimension scores to isolate their relative contributions:

$$\text{visual vividness}_i = \beta_0 + \beta_1 \cdot \text{head}_{i,} + \beta_2 \cdot \text{hand}_i + \beta_3 \cdot \text{mouth}_i + \beta_4 \cdot \text{foot}_i + \beta_5 \cdot \text{torso} + \epsilon_i$$

All predictors were z-scored prior to model fitting. GLM analyses were conducted in *R* using the *stats* package (R Core Team, 2023).

**Controlling for response length via mediation**. Initial analysis revealed that description length was a significant positive predictor of vividness ratings ($\beta = .01$, $SE = .00$, $t = 6.80$, $p < .001$), indicating that participants with more vivid imagery provided longer descriptions. To account for this potential confounding influence, we conducted a series of causal mediation analyses using the *mediation* package in *R* (Tingley et al., 2014). Specifically, for each set of predictors—(1) composite perceptual and action strength, (2) the six perceptual modality scores, and (3) the five motor modality scores—we assessed whether the effect of sensorimotor richness on visual vividness was mediated by description length.

Each mediation analysis included two models: a *mediator model* predicting standardized description length from the sensorimotor predictors of interest, and an *outcome model* predicting visual vividness from both the sensorimotor predictors and standardized description length.

Mediator model (example for a single predictor):

$$\text{description length}_i = \gamma, + \gamma, \cdot \text{sensorimotor predictor}_i + \eta_i$$

Outcome model:

$$\text{visual vividness}_i = \beta_0 + \beta_1 \cdot \text{sensorimotor predictor}_i + \beta_2 \cdot \text{description length}_i + \epsilon_i$$

We estimated the average direct effect (ADE), average causal mediation effect (ACME), total effect, and proportion mediated for each predictor using 5,000 nonparametric bootstrap simulations with percentile-based confidence intervals.

## Results

### Unsupervised topic modeling identifies systematic themes across the imagery spectrum

To map the latent content structure of participants' hallucination descriptions, we used topic modeling. This unsupervised approach identifies patterns of word co-occurrence and groups semantically similar descriptions together, allowing us to reveal the types of experiences participants reported without imposing predetermined categories (Blei et al., 2003). Applying this method to the corpus yielded 28 distinct experiential themes, each reflecting a coherent hallucination type (e.g., geometric patterns, faces, tunnels, stars; Table 1).

The discovered topics encompassed a diverse range of hallucinations—from basic visual elements (e.g., lines, dots, color flashes), to scenery (e.g., forests and skylines), and socially-relevant objects (e.g., faces and figures) (Fig. 6, Table 1). Representative quotes from descriptions associated with each topic are in Table 1. Each row shows a free-text quote drawn from participant descriptions of their Ganzflicker-induced hallucinations, alongside the topic with which it was most strongly associated. For instance, the topic spirals & rotation included descriptions like "everything constantly spinning," while the topic Space was associated with reports like "flying through a space station." These examples show that our unsupervised model successfully captured linguistically distinct yet conceptually similar experiences, ranging from motion and geometric percepts to naturalistic scenes and affective content.

Semantic structure in the 2D Uniform Manifold Approximation and Projection (UMAP; McInnes et al., 2018) embedding showed clearly delineated clusters, where spatial proximity reflects thematic similarity (Fig. 6). Each point represents a sentence from a participant's hallucination description, embedded into a shared semantic space and colored by its assigned topic. The visualization revealed several broad regions: geometric and motion-related content (e.g., patterns & shapes, fractals, spirals & rotation, lines, dots, movement) formed one area; naturalistic and environmental imagery (e.g., forest & trees, water, flowers, space, city skyline, tunnels) formed another; and metacognitive domains (e.g., sensation duration and hallucination absence) appeared in a more separate region. Together, these analyses indicate that hallucination descriptions organize into coherent semantic domains with consistent relationships among topics. The resulting topic structure depicted in Figure 6 offers a data-driven foundation to examine whether hallucination content varies systematically across individuals with different imagery phenotypes.

**Table 1 Representative hallucination descriptions by topic.**

| Human labels | Model labels | Representative descriptions |
|---|---|---|
| Experiential Descriptions | Trippy Experience | *"it was like watching a trippy video of colours and sounds"* |
| Hallucination Absence | Minimal Content | *"I didn't see anything substantial though I thought I would"* |
| Noise | White Noise | *"the white noise and screen together gave me an uneasy feeling but nothing noteworthy just a if this was a horror movie this would be the part where someone dies feeling"* |
| Faces & Figures | Human Faces | *"a potato that turned into an old mans face and then some alien scenery that was pulsing"* |
| Unpleasant Images | Subjective Experience | *"it was not an enjoyable experience but i felt no overwhelming discomfort else i would have stopped the experiment"* |
| Eyes | Eye Patterns | *"after several minutes i saw a left eye sketchlike which rotated and disappeared then reappeared several times"* |
| Sensation Duration | Duration Events | *"Images only lasted for a few seconds each"* |
| Images | Images Movement | *"after a few seconds of this i usually would have to blink and the images would be gone"* |
| Gaze | Eye Variations | *"as soon as i tried to look at them or focus on them they would disappear"* |
| Webs | Spider Webs | *"patterns often in the form of spider webs"* |
| Dots | Dots Pattern | "tiny dots of other colours as part of the pattern like smaller than pixels and so many that they almost made up other colours" |
| Colors | Colors & Patterns | *"the black ended up shifting between a greenish black and deep blue"* |
| Movement | Visual Distortion | *"mostly just different colors bouncing all around the screen in different shapes"* |
| Flashes | Flashing Lights | *"blue pink yellow flashes but mostly a continuous pink flash"* |
| Lines | Lines & Patterns | *“thick horizontal and vertical lines with fuzzy edges that intersected at the center of my field of view”* |
| Patterns & Shapes | Geometric Patterns | *“i saw a bunch of shapes like squares circles lines”* |
| Fractals | Fractals & Patterns | *“like a kaleidoscope but much more faint and ever changing”* |
| Spirals & Rotation | Spiral Rotation | *“everything constantly spinning”* |
| Stars & Galaxies | Stars & Galaxies | *“i saw purple dots that were circling around like stars in a galaxy also with spiral arms etc”* |
| Flowers | Flower Imagery | *“perceived i was looking at a field of tulips and daisys and flowers as if through a crumpling red fold of cellophane”* |
| Fire | Fire Imagery | *“i saw a starburst firework style shape occasionally and a couple of little sparks like a sparkler”* |
| City Skyline | Cityscape Views | *“reminded me of looking at a bright light with my eyes closed”* |
| Forest & Trees | Forest Scenery | *“On the far edge and wrapping around the shore was a treeline that consisted of birch like trees tall and thinner with plenty of canopy but were not birches and some juniper trees it was accompanied with average pond foliage like cattails and reeds and such”* |
| Water | Ocean Reflections | *“it wasnt realistically colored or anything just felt like i was standing on the edge of a lake looking at something in the middle of it”* |
| Space | Space Travel | *“flying through a space station”* |
| Tunnels | Tunnel Experience | *“a circle would appear then it turned into a long tunnel”* |
| Hallways & Doors | Hallway Passage | *“also a very long dark corridor with light at the end it was moving around on the vertical axis”* |
| Butterflies | Butterfly Shape | *“first was like a butterfly outline with neon lines”* |

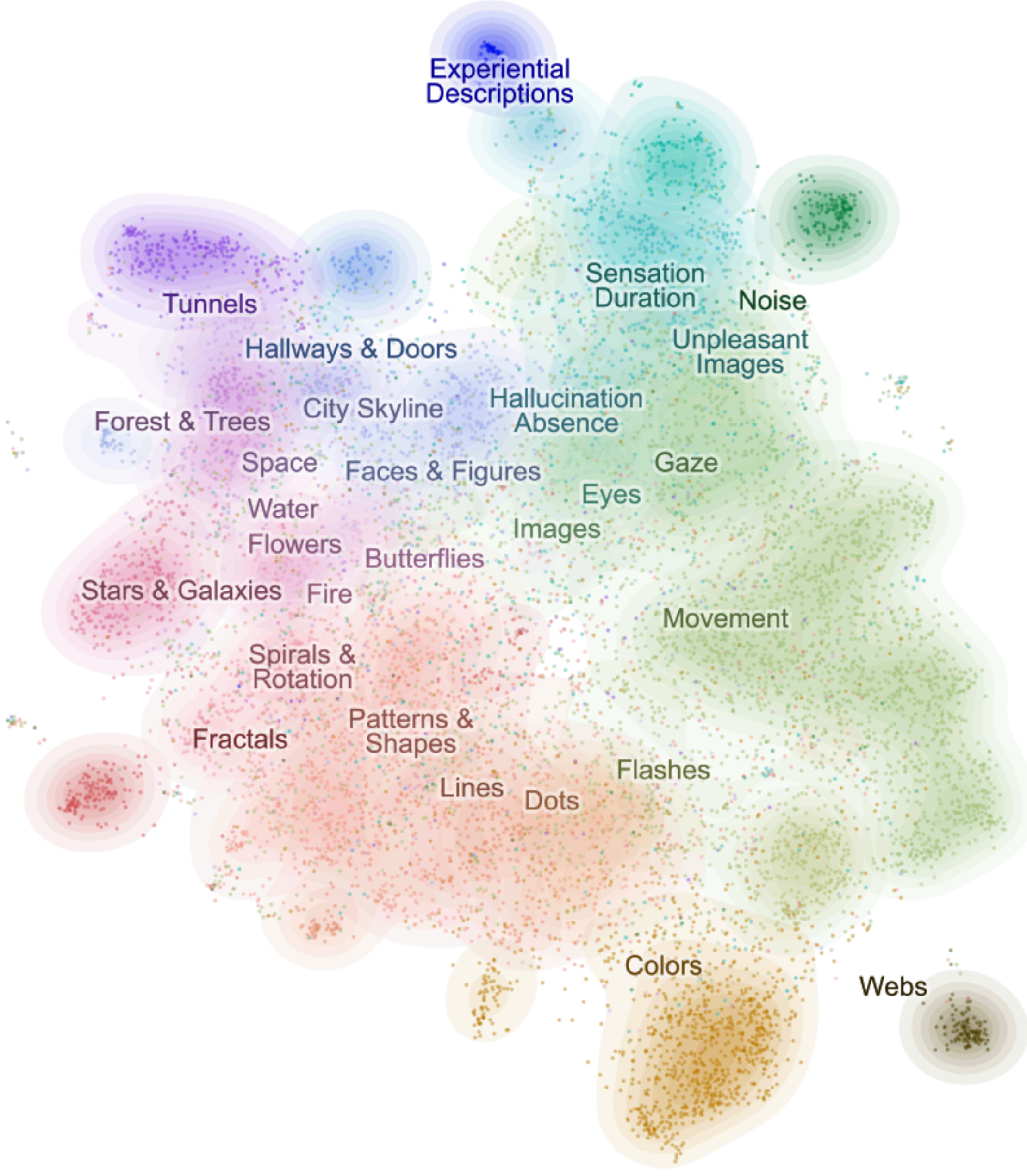

**Figure 6. Topic representations and hierarchical relationships among hallucinations.** UMAP projection of sentence embeddings colored by topic labels. Each point represents a sentence from a participant's hallucination description embedded in a shared semantic space. Clusters correspond to semantically coherent hallucination types, ranging from simple geometric content to naturalistic scenes and more abstract descriptions.

We hypothesized that participants with stronger imagery would describe more complex, naturalistic hallucination content, whereas weaker imagers would report simpler visual distortions and geometric forms. We observed systematic content differences consistent with this prediction: strong imagers more often reported structured, semantically meaningful content (e.g., faces, hallways, natural environments), while weak imagers tended to describe simpler features such as geometric patterns, visual distortions, and flashes (Fig. 7).

To test this quantitatively, we first modeled participants' continuous visual imagery vividness scores using Lasso regression (Tibshirani, 1996). Given any particular hallucination description, the topic model output provides a probabilistic profile across all 28 topics, reflecting the estimated content composition of that report. We used these topic probabilities as features in our models to predict imagery scores. Topic probabilities explained approximately 5% of the variance in vividness scores ($R^2 = .05$). From the initial set of 28 topics, 25 were retained as non-zero predictors (see Fig. S2). Given the noisy nature of free-text reports and the multifaceted character of subjective experience, this effect reflects meaningful structure in the data despite its modest size. As human psychology and behavior is inherently complex, small but reliable effects in psychological research are both expected and interpretable (Funder & Ozer, 2019). Positive predictors included naturalistic or structured content—faces and figures ($\beta = 0.29$), hallways and doors ($\beta = 0.15$), stars ($\beta = 0.14$), city skylines ($\beta = 0.14$), and forest imagery ($\beta = 0.13$). Negative predictors consisted mainly of simple perceptual content, including movement ($\beta = -0.23$), flashes ($\beta = -0.17$), and lines ($\beta = -0.12$). These results indicate that richer, more naturalistic descriptions are associated with higher imagery vividness, while lower vividness is linked to simpler perceptual experiences.

To characterize the hallucination content profiles of distinct imagery phenotypes, we categorized participants into three vividness groups: weak (0–3), moderate (4–7), and strong (8–10). This binning followed conventions in the imagery literature distinguishing aphantasia, typical imagery, and hyperphantasia (Zeman, 2024). For each group, we computed the average topic probability across participants (Fig. S3). To assess whether these profiles were discriminable, we then trained Lasso-regularized classifiers to predict group membership from topic probability features. We trained three separate one-vs-rest classifiers, where each classifier learned to identify one imagery group versus all others, revealing which topics are most diagnostic of each imagery phenotype. The classifiers were evaluated using the F1 score—the harmonic mean of precision and recall—which is well-suited to imbalanced classification tasks and helps assess how well models identify the minority class (Saito & Rehmsmeier, 2015). For statistical validation, we conducted permutation tests (Nichols & Holmes, 2002) in which group labels were randomly shuffled and each model retrained 1,000 times. The weak-imagery classifier showed significant above-chance performance (F1 = 0.54, $p < .001$), indicating that those with the least vivid imagery can be distinguished from other participants based on their hallucination content. Classifiers for the moderate (F1 = 0.45) and strong groups (F1 = 0.41) did not exceed chance ($p > .50$), likely reflecting similarity between these two groups (Fig. S3).

To interpret classifier outputs, we examined bootstrapped Lasso coefficients for the topics and retained only those that appeared in at least 60% of 1,000 bootstrapped models (Fig. 7). The weak-imagery profile was sparsely defined: only 8 of 28 topics were retained, mostly reflecting the absence of naturalistic content. Positive predictors for weak imagery included simple perceptual features (e.g., flashes, movement). In contrast, all 28 topics survived regularization and bootstrapping for the moderate and strong groups, and their strongest predictors overlapped with those from the continuous model. Notably, topics absent in weak-imagery descriptions (e.g., faces, natural scenes) were positive predictors of strong imagery, whereas topics characteristic of weak imagery were negative predictors of strong imagery (Fig. 7).

Together, these findings suggest that free-text descriptions of hallucinations contain discriminative signals related to individual imagery phenotypes. The semantic content of these descriptions offers a high-dimensional, quantifiable window into how internal visual experiences differ across the imagery spectrum.

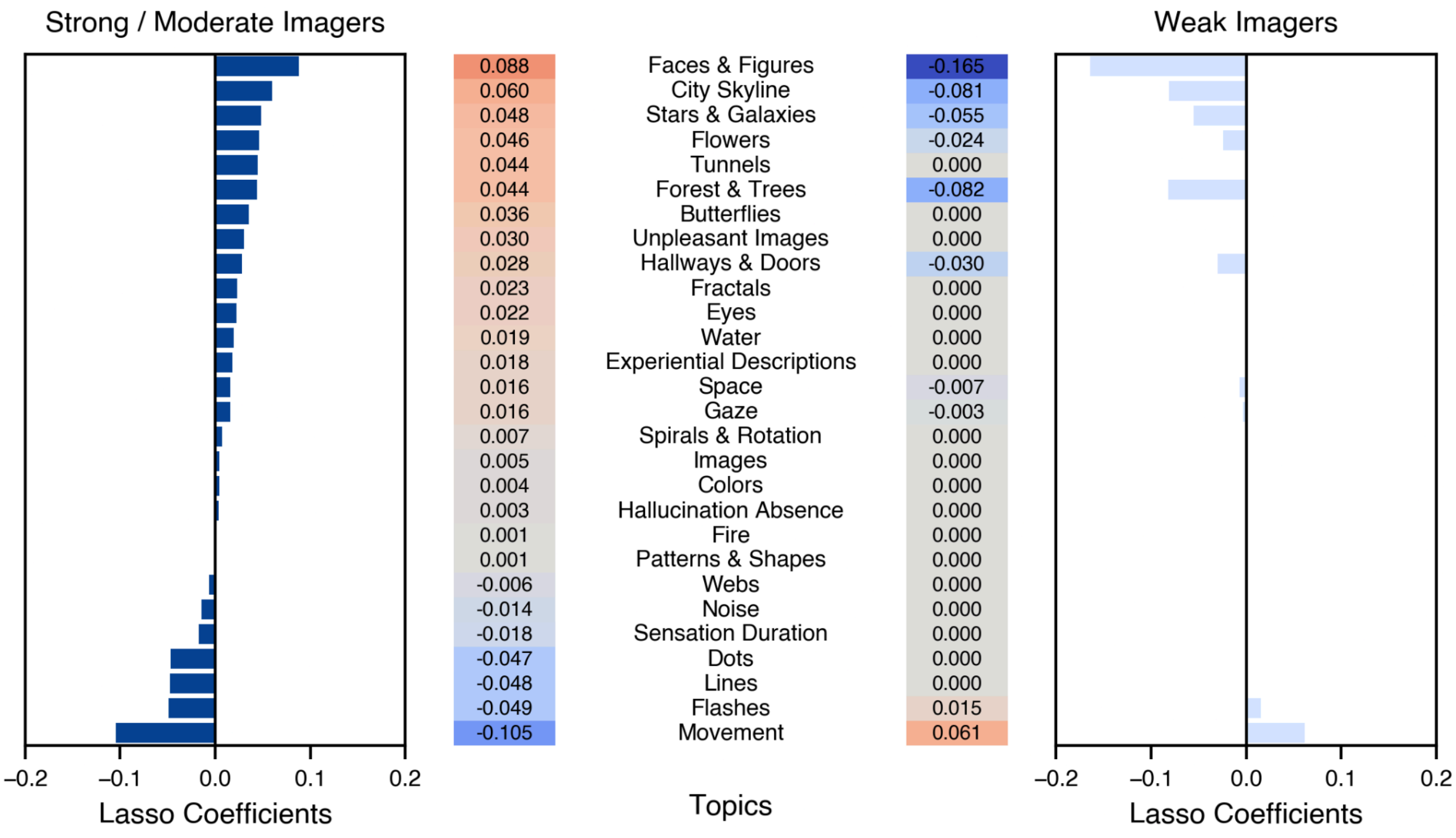

**Figure 7. Hallucination content patterns distinguishing weak from moderate / strong imagery groups.** Lasso regression coefficients averaged across the moderate and strong imagery groups and contrasted with the weak group. The plot shows only topics that survived regularization and appeared in at least 60% of 1,000 bootstrapped regressions. Colors indicate coefficient direction and stability, with positive values reflecting topics more prevalent in the moderate/strong imagery group and negative values reflecting topics more common in the weak group.

### Human-assessed sensorimotor content of descriptions varies with imagery vividness

We expected that participants with stronger imagery vividness would use language with stronger sensorimotor associations when describing their hallucinations, particularly in visual and motor dimensions, reflecting richer internal simulation.

To quantify this content, we obtained Lancaster scores for each word in participants' descriptions across six perceptual modalities, five motor dimensions, and two composite measures, then averaged these scores within each participant to create 13 participant-level sensorimotor features (see Fig. 8a for analysis pipeline). Across the dataset, visual strength was the most prominent perceptual modality, reflected in its relatively high average score (M = 3.57) and substantial individual variability (range = 3.87), followed by haptic (M = 1.32) and auditory (M = 1.11) dimensions. In the motor domain, head- (M = 2.66) and hand-related (M = 1.56) concepts were most frequently represented. The composite measures of perceptual and action strength also showed high mean scores and variability, supporting their interpretability as global indices of sensorimotor richness. Full score distributions and descriptive statistics are reported in Table S1 and Fig. S4.

We first modeled imagery vividness using composite perceptual and motor (action) strength, controlling for description length. We entered the composite perceptual and motor (action) strength scores from the Lancaster Sensorimotor norms into a GLM, along with standardized description length as a covariate. Both predictors were significant: higher perceptual strength was associated with greater imagery vividness ($\beta = 0.35$, SE = 0.05, $t = 6.54$, $p < .001$), as was higher motor strength ($\beta = 0.21$, SE = 0.05, $t = 4.17$, $p < .001$) (Fig. 8b). Description length was also a significant positive predictor ($\beta = 0.42$, SE = 0.05, $t = 8.39$, $p < .001$), consistent with the idea that more vivid experiences elicit richer verbal responses.

To determine whether the effects of sensorimotor content were confounded or mediated by description length, we conducted causal mediation analyses with 5,000 bootstrap simulations. For perceptual strength, the total effect ($\beta = 0.23$, $p < .001$) was composed of a positive direct effect (ADE = 0.35, $p < .001$) and a significant *negative* indirect effect via description length (ACME = –0.12, $p < .001$), indicating a suppressive mediation. That is, the full influence of perceptual content was actually underestimated when not controlling for description length, with over half the total effect (–54%) masked by its association with longer responses. In contrast, action strength showed a small but significant positive mediation effect (ACME = 0.04, $p < .001$), contributing to a total effect of 0.25 ($p < .001$), with 16% of this effect attributable to longer responses. These results suggest that both perceptual and action content are independently predictive of visual imagery vividness.

To identify which specific sensory modalities contributed to the observed effect of perceptual strength on visual imagery vividness, we fit a generalized linear model including all six Lancaster perceptual dimensions—visual, auditory, gustatory, olfactory, haptic, and interoceptive—as simultaneous predictors (see Fig. S5 for individual dimension trends). Description length was included as a covariate to control for potential verbosity differences.

The results revealed that five of the six perceptual modalities reliably predicted imagery vividness (Fig. 8c, left). Stronger visual content in hallucination descriptions was associated with higher vividness ratings (β = 0.40, $p < .001$), as was greater haptic (β= 0.32, $p < .001$), olfactory (β = 0.36, $p < .001$), and auditory content (β = 0.18, $p = .003$). Interestingly, gustatory strength was negatively associated with vividness (β = –0.28, $p < .001$), while interoceptive content showed a positive but non-significant effect (β = 0.12, $p = .093$).

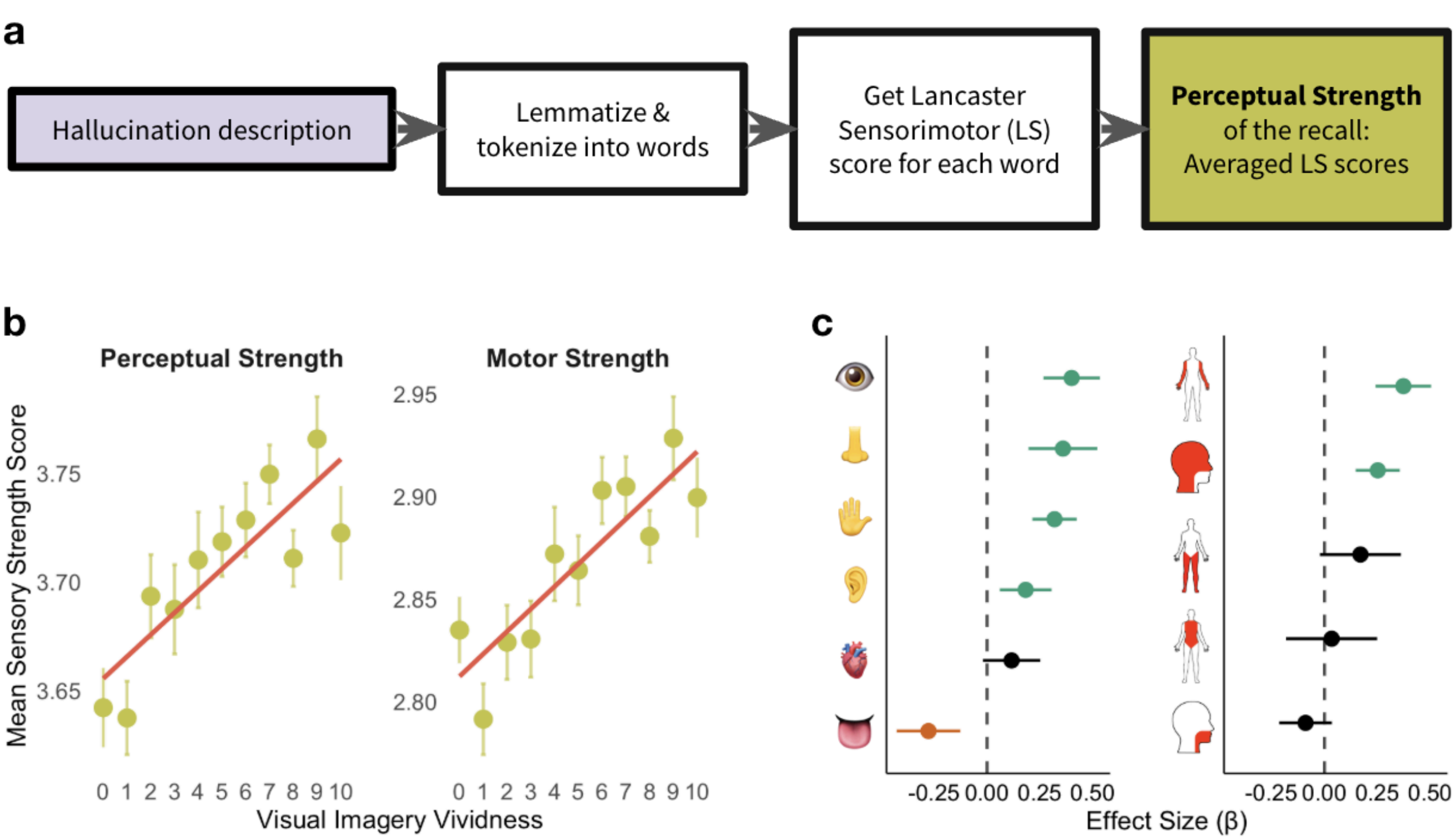

**Figure 8. Lancaster Sensorimotor analysis pipeline and results. a,** Analysis pipeline showing how hallucination descriptions were processed to extract sensorimotor content using the Lancaster Sensorimotor Norms. Text was lemmatized and tokenized, each word received Lancaster Sensorimotor scores, and perceptual strength was calculated by averaging LS scores of individual words in each description. **b,** Relationship between visual imagery vividness and composite sensorimotor dimensions. Both perceptual strength (left) and motor strength (right) showed significant positive associations with self-reported imagery vividness. Points represent mean scores ± standard error for each vividness level. **c,** Forest plots showing effect sizes (β coefficients) from generalized linear models predicting imagery vividness from individual sensorimotor dimensions, controlling for description length. Left panel shows perceptual modalities with icons representing visual (eye), olfactory (nose), haptic (hand), auditory (ear), interoceptive (heart), and gustatory (tongue) dimensions. Right panel shows motor dimensions with body part icons for head, hand, mouth, torso, and foot. Colored points indicate significant effects (teal = positive, orange = negative), black points indicate non-significant effects. Error bars represent 95% confidence intervals.

To account for description length, we ran mediation analyses for each modality. These analyses confirmed that the core effects remained robust even after controlling for verbosity. Specifically, the direct effect of visual strength remained strong and positive ($ADE$ = 0.40, $p < .001$), even though the total effect was slightly suppressed by a significant negative mediation via description length ($ACME$ = –0.10). This indicates that participants with more vivid imagery did not merely write longer descriptions; rather, the visual content of those descriptions was independently predictive of vividness. Haptic and olfactory strength also retained significant direct effects after mediation (*ADEs* = 0.32 and 0.36, respectively; both $p < .001$), suggesting that these less commonly studied sensory modalities were meaningfully tied to self-reported individual differences in visual imagery. Although olfactory strength survived mediation analysis, its low mean and restricted range suggest that it should be interpreted with caution, as it may reflect isolated rather than widespread patterns of olfactory content. Auditory content also remained significant (*ADE* = 0.18, $p = .003$), though its mediated component was small but positive (*ACME* = 0.022). The effect of auditory content on imagery vividness is mostly direct and robust, although there was a small, significant indirect effect of description length. The effect of interoceptive content remained non-significant after mediation (*ADE* = 0.12, $p = .10$; CI includes 0), indicating no additional influence of verbosity. While gustatory strength showed a significant negative coefficient in the regression model, its mediation results and real-data trend (Fig. S5) suggest interpretational caution. The model returned a negative direct effect (*ADE* = –0.28, $p < .001$), but raw data binned by vividness showed a slight upward trend in gustatory content, likely driven by sparse outliers. This dimension also had the lowest mean (0.30) and smallest variance (0.035) among all predictors, suggesting that gustatory language was rarely present and potentially contributed noise. Given this, we refrain from interpreting its effect as meaningful.

Together, these findings suggest that visual imagery vividness is reflected not only in explicitly visual language but also in modality-rich descriptions involving haptic, olfactory, and auditory references in addition to action-related language. The robustness of these effects after controlling for verbosity strengthens the claim that these are not mere artifacts of expressive ability or fluency.

To determine which body-related dimensions accounted for the effect of action strength on visual imagery vividness, we next modeled each of the five Lancaster motor dimensions—head, hand, mouth, torso, and foot strength—as simultaneous predictors of vividness (Fig. S5 shows individual trends for each motor dimension). Description length was included as a covariate. The model revealed significant positive effects for hand strength ($\beta = 0.37$, $p < .001$) and head strength ($\beta = 0.25$, $p < .001$), suggesting that vivid imagers were more likely to describe content referencing the head or hands (e.g., “looking around,” “reaching,” “hands in front of me”). The effects of foot, mouth, and torso strength were not statistically significant ($p$s > .05) (Fig. 8c, right).

We followed up this model with separate causal mediation analyses to account for the role of description length. As in earlier models, mediation analyses were conducted for each motor dimension using 5,000 bootstrap simulations. Both hand strength and head strength remained robust predictors after controlling for length. For hand strength, the direct effect (ADE) remained significant (ADE = 0.37, $p < .001$), while the indirect effect (ACME = –0.056, $p < .001$) indicated a suppressive relationship with length: longer responses slightly diluted the observed association, but did not eliminate it. A similar pattern held for head strength (ADE = 0.25, $p$ <

.001; ACME = –0.031, $p < .001$). In contrast, the effects of foot, mouth, and torso strength did not survive mediation analysis. Foot strength showed a small positive total effect (0.17), but neither the direct nor indirect paths reached significance ($p$ = .07 and .78, respectively). Torso strength had a positive indirect effect (ACME = 0.031, $p < .001$), but its direct and total effects were non-significant, and its mediation estimate was highly unstable (Prop. Mediated = 0.48, 95% CI = –3.77 to 4.10). Mouth strength showed a suppressive pattern (Prop. Mediated = –1.65), but neither its direct nor total effect was significant ($p$ = .16), and the instability of this estimate again likely reflects noise due to sparse representation.

These results highlight head- and hand-related content as the most consistent motor predictors of imagery vividness. Their predictive power held even when accounting for description length and shared variance with other motor dimensions. This pattern suggests that vivid internal experiences are more likely to include references to sensorimotor processes involving head movements and hand interactions.

Prior work shows that women report higher visual vividness scores than men on the VVIQ (Richardson, 1995; Aydin, 2020) and self-reported vividness declines across the lifespan (Gulyás et al., 2022). To evaluate whether these factors influenced imagery ratings in our sample, we regressed vividness scores on age and gender. Age was not a significant predictor ($\beta = -0.0002$, $p = .95$), but consistent with the prior literature, women reported reliably higher vividness than men ($\beta = -0.53$, $p < .001$). To ensure that the main effects reported above were not confounded by gender differences, we re-ran all Lancaster sensorimotor models including gender as a covariate. All sensorimotor predictors that were significant in the primary models remained significant with minimal change in effect size (Fig. S6).

# Discussion

## Overview

Through computational analysis of over 4,000 phenomenological reports, we characterized the content and structure of Ganzflicker-induced hallucinations and examined how this content varies across the visual imagery spectrum. Results from our complementary analytical approaches—topic modeling and sensorimotor norms—converged to reveal systematic differences in hallucination phenomenology consistent with the layered model of visual imagery. Below, we situate our findings within the broader literature on experimentally-induced hallucinations and altered states of consciousness (ASC), and discuss what we learned about individual differences in visual imagery.

### Phenomenological profile of Ganzflicker-induced hallucinations

Our topic modeling identified 28 distinct experiential themes, providing a detailed characterization of Ganzflicker-induced hallucination phenomenology. This granular phenomenological mapping extends prior work on flicker-induced experiences, which has typically relied on structured questionnaires or broad categorical distinctions (Hewitt et al., 2025; Bartossek et al., 2021). Consistent with decades of research on light-induced hallucinations (Klüver, 1926; Bressloff et al., 2001, 2002; Hewitt et al., 2025), simple geometric content was among the most frequently reported experiences. Our participants described classical "form constants" (Klüver, 1926): spirals, tunnels, and cobwebs that recur across diverse altered states and induction methods, from Ganzfeld (Pistolas & Wagemans, 2025; Wackermann et al., 2008) to stroboscopic light stimulation (Hewitt et al., 2025; Bartossek et al., 2021). However, our analysis also documented complex, semantically meaningful content: faces, human figures, city skylines, forest scenery, stars and galaxies—naturalistic imagery that has previously been reported in flicker paradigms (Allefeld et al., 2011; Beauté et al., 2025; Königsmark et al., 2021; Schwartzman et al., 2019; Shenyan et al., 2024), but not characterized at this level of detail.

The hierarchical clustering of these 28 topics (Fig. S1) revealed three experiential domains that mirror the visual processing hierarchy: simple geometric content (e.g., lines, flashes, patterns, dots, colors), naturalistic scenes (e.g., skylines, forests, water, butterflies), and a cluster of social (faces, figures, eyes) and metacognitive (experiential descriptions, sensation duration) topics. This data-driven organization aligns with theoretical frameworks proposing that hallucinations evolve from geometric forms to structured experiences as higher-order visual areas become engaged (ffytche, 2008; Reeder et al., 2024).

We also examined whether Ganzflicker induced alterations in consciousness beyond visual hallucinations, following the taxonomy of ASC features introduced by Fort et al. (2025). While our results primarily implicated the Perception and Imagery dimensions, participants also reported experiences that may relate to other ASC features: feelings of being "transported to another place," "dreamlike state," and "lost sense of space" share phenomenological overlap with the Self-Boundary feature (which includes depersonalization and derealization experiences), while "lost a sense of time" aligns with the Time Sense feature. Participants also reported Emotion changes ranging from enjoyable to unpleasant. These findings situate Ganzflicker within the broader ASC literature, demonstrating that, like other non-pharmacological induction methods (Berkovich-Ohana et al., 2013; Kübel et al., 2021), it produces alterations extending beyond perception to encompass spatial disorientation, dreamlike absorption, and emotional engagement.

Importantly, experienced ASCs varied systematically across the imagery spectrum. Vivid imagers were significantly more likely to report feeling Self-Boundary alterations, including feelings of being transported, dreamlike states, and spatial disorientation (Fig. S7). By contrast, weak imagers were more likely to report no changes in consciousness. Emotional experiences also varied across individuals: vivid imagers rated their experience as more enjoyable (Fig. S8), even while describing more affectively intense visual content (e.g., "unpleasant images"). This finding suggests that vivid hallucinations may be experienced as engaging even when their content is disturbing, and fits with the claim that the Emotion feature interacts with but is conceptually distinct from perceptual content (Fort et al., 2025). Imagery was not related to the alterations in time perception, suggesting that Time Sense might be a relatively stable and broadly shared feature of Gazflicker experience, invariant across the imagery spectrum. Beyond these broader experiential differences, hallucination content itself varied systematically with imagery capacity.

**Hallucinated content maps onto the visual processing hierarchy across the imagery spectrum**

Imagery research has traditionally treated vividness as a single continuum, capturing how clearly or intensely mental images are experienced. Our findings suggest that individuals differ not just in the intensity of their imagery, but the level of representational complexity they can achieve. Accordingly, these findings are consistent with the layered model of visual imagery outlined in the introduction (Fig. 2). Inspired by neuroimaging evidence that aphantasic individuals retain early visual representations during imagery (Cabbai et al., 2024; Chang et al., 2025) this model holds that all individuals represent basic visual features in the early visual cortex. Further, since aphantasics show reduced functional connectivity between frontoparietal control networks and higher-order visual regions (Liu et al., 2025; Milton et al., 2021; Spagna et al., 2021), the layered model suggests stronger imagers more effectively engage higher-order regions to integrate basic visual features into coherent, semantically meaningful representations.

If imagery vividness is a scalar intensity parameter affecting all visual processing equally, we would expect proportional scaling of both simple and complex content across the imagery spectrum. Instead, we observed a qualitative shift: weak imagers described the building blocks of visual experience, while strong imagers reported content requiring top-down integration across the full cortical hierarchy. We formalize this mapping in Table 2, which links hallucination topics to their predicted processing levels based on their association with imagery vividness.

Even individuals reporting the complete absence of imagery described seeing something during Ganzflicker—but what they saw differed qualitatively from what stronger imagers reported. Weak imagers predominantly described simple visual experiences like geometric forms: lines, dots, flashes, and movement—content consistent with accounts of the early visual cortex involvement during hallucinations, like form constants arising from V1/V2 dynamics (Bressloff et al., 2002; Burke, 2002; Ermentrout & Cowan, 1979; Rule et al., 2011) and motion-selective intermediate areas like V3/MT (Amaya et al., 2023, 2025). Strong imagers, by contrast, reported complex, naturalistic, and socially relevant content: faces and figures, city skylines, forests, butterflies, and immersive spatial experiences. This type of content engages later extrastriate and ventral temporal regions that represent more complex and multi-dimensional stimuli (Epstein & Kanwisher, 1998; Kanwisher et al., 1997).

**Table 2. Hallucination content mapped onto visual processing hierarchy across the imagery spectrum.**

| Processing level | Area | Weak imagers / Aphantasia | Moderate-Strong Imagers |
|---|---|---|---|
| Visual primitives | V1/V2 | lines, dots, flashes | lines, dots, flashes |
| Intermediate | V3/V4, MT/V5 | movement, colors, | movement, colors, patterns and shapes, spirals and rotation, tunnels, webs |
| Semantically meaningful objects | IT | — | butterflies, flowers, stars and galaxies, faces and figures, eyes, city skyline, forests and trees, hallways and doors, space |
| Control / Integration | Frontoparietal, Insula | — | experiential descriptions, sensation duration |

Neural accounts of flicker-induced hallucinations also support our proposed link between individual differences in visual imagery and the extent of hierarchical interaction in the visual system. For example, computational models propose that simple geometric hallucinations arise from dynamical interactions within early visual cortex, where local excitatory and long-range inhibitory connections generate repeating patterns that resemble Klüver's form constants (Bressloff et al., 2002; Ermentrout & Cowan, 1979; Rule et al., 2011). Complex naturalistic content, by contrast, requires engagement of higher-order visual areas (Shenyan et al., 2024; Stoliker et al., 2025). In addition to cortical hierarchical models, neural accounts have also highlighted a potential role for thalamocortical interactions in flicker-induced hallucinations (see Hewitt et al., 2025 for a review).

In fact, more effective thalamocortical engagement of higher-order visual areas could explain the individual differences in hallucination complexity observed in the present study. The thalamus acts as a central hub regulating cortical excitability and integrating information across networks (Hwang et al., 2017; Shine, 2021), with ventral projections targeting supragranular cortical layers implicated in conscious experience (Phillips et al., 2019; Suzuki & Larkum, 2020). Just as enhanced top-down connectivity has been shown to be relevant for hyperphantasics' voluntary imagery (Milton et al., 2021), fMRI studies reveal that periodic flicker stimulation produces thalamocortical hyperconnectivity between ventral thalamic nuclei and higher-order visual cortices, with connectivity strength correlating with subjective experience of hallucination intensity (Amaya et al., 2023, 2025). Future studies could test whether there is a relationship between imagery capacity, thalamocortical connectivity, and the complexity of flicker-induced hallucinations, as well as the complexity of other experiences that rely on the generative capacity of the visual system.

**Sensorimotor content of hallucination descriptions may reflect multisensory simulation**

To complement computational analyses with human-derived measures, we turned to the Lancaster Sensorimotor Norms to examine the perceptual and action-related content of participants' descriptions using crowdsourced judgments. We found that descriptions from individuals with higher imagery vividness contained more words with strong visual, haptic, and auditory content, as well as head- and hand-related motor associations. Notably, stronger imagers used richer language not only in the visual domain but also for other perceptual content (see Fig. 8c), consistent with evidence that imagery differences often extend beyond vision, with many aphantasics reporting weak imagery across sensory modalities (Zeman et al., 2020). This suggests that what we measure as "visual" imagery vividness may index broader variation in multisensory simulation capacity (Dance et al., 2021). In short, strong imagers used language with richer perceptual connotations, suggestive of more grounded and content-dense internal simulations.

Potentially at odds with predictive-coding accounts proposing that interoception and agency play a central role in imagery generation (Silvanto, 2025; Silvanto & Nagai, 2025), we failed to find a relationship between interoceptive language and imagery strength. However, the Lancaster norms quantify explicit interoceptive language, whereas the relevant theoretical accounts emphasize implicit mechanisms like agency or bodily prediction. These may not surface in verbal descriptions when participants are instructed to report what they saw. When directly asked about their conscious experience, however, stronger imagers were more likely to report spatial disorientation and dreamlike absorption, and to evaluate the experience as emotionally enjoyable compared to those with weak visual imagery(Figs. S7 and S8). This confirms that embodied and affective experience do vary with imagery vividness, even when not captured by explicit language norms.

The most robust predictors of individuals' visual imagery vividness were visual, head- and hand-related language, consistent with our layered model, in which stronger imagers more effectively engage high-level perceptual systems that support complex, socially meaningful internal content. Faces and hands are among the most socially and evolutionarily significant concepts for humans (Bracci et al., 2010; Lacruz et al., 2019), each with its own dedicated neural circuitry (Bracci et al., 2010; Kanwisher et al., 1997). When higher-order areas are recruited during complex hallucinations, these strongest "defaults" may surface first. As individuals with aphantasia show reduced activation of the fusiform gyrus during imagery tasks (Liu et al., 2025), stronger imagers might more effectively recruit specialized regions such as the fusiform face area, during Ganzflicker-induced hallucinations, resulting in experiencing richer and more complex visual content, as well as possibly richer multisensory associations.

## Limitations

Our study drew on a large online sample (~4,000 participants), allowing us to capture rare individual differences (e.g., aphantasia and hyperphantasia) and to examine Ganzflicker experiences across diverse contexts. However, online studies come with inherent constraints. For example, we could not directly control participants' viewing environments. Although all participants viewed the same 7.5 Hz red–black flicker stimulus, variations in the properties of their monitors (e.g., luminance, color saturation, timing stability) and viewing conditions (e.g., lighting, distractions, use of white noise) may have introduced heterogeneity into their perceptual responses. Without standardized conditions, the data could in principle represent a heterogeneous set of experiences rather than differences related to participants' position on the imagery spectrum.

However, several factors support the interpretation that our findings reflect genuine Ganzflicker phenomenology. Prior analyses of the impact of environmental variables in this dataset, including white noise use, lighting conditions, or display medium, failed to reveal significant effects on reported hallucination content (see Reeder, 2022 for details; see also Königsmark et al., 2021). Further, the hallucination content documented here aligns with that reported in prior Ganzflicker studies using different samples and setups (Königsmark et al., 2021), including one conducted in a controlled laboratory environment (Shenyan et al., 2024). Nevertheless, online data collection on its own cannot fully rule out the contribution of uncontrolled environmental factors, and future work with larger controlled laboratory samples would strengthen confidence in these findings.

Another limitation of our study relates to participant recruitment via a public article about Ganzflicker. It seems likely that this may have attracted individuals already interested in hallucination phenomena as well as those at the extremes of the imagery spectrum, in keeping with the higher-than-typical proportion of self-reported aphantasia in our sample. This raises the possibility that observed results reflect demand or expectancy effects, with those interested in hallucinations eager to report ASCs and those who identify as aphantasic eager to produce descriptions reflecting their expectations rather than their actual experiences. However, Ganzflicker-induced hallucinations reliably emerge in laboratory settings without self-selection (Shenyan et al., 2024). Further, in the present study, even participants reporting no visual imagery described hallucinations, arguing against expectancy alone as an explanation (Reeder, 2022). Finally, hallucination content changed predictably with imagery vividness—a pattern inconsistent with simple demand effects.

Yet another limitation of the present study is our use of participants' verbal descriptions as an index of their subjective visual experiences. Although language provides a powerful window into the content of participants' hallucinations, their descriptions necessarily conflate phenomenology with narrative expression. Individuals may differ not only in what they experience but also in how they verbalize those experiences. Notably, participants with more vivid imagery provided longer descriptions, raising the possibility that verbosity rather than experiential richness drives some effects. To address this, our analyses of Lancaster Sensorimotor Norms incorporated measures to control for description length, finding that vividness effects remained significant. Corroborating evidence also comes from closed-ended measures: participants with higher imagery vividness reported greater alterations in consciousness and more positive affect during Ganzflicker stimulation on questions that did not require generating linguistic descriptions (Figs. S7, S8). That said, verbal reports alone cannot adjudicate between differences in what people experience and differences in how they describe itFuture work incorporating objective measures such as EEG, eye tracking, and physiological markers will be critical for directly assessing perceptual dynamics during Ganzflicker stimulation.

## Conclusions

Together, these findings offer a rich characterization of Ganzflicker-induced experience, from the diversity of visual hallucinations to broader alterations in consciousness, and reveal how both vary systematically across the visual imagery spectrum. Overall, results align with the theory of divergent predictive perception (Reeder et al., 2024), which proposes that individuals across the imagery spectrum differ in how strongly top-down predictions shape perceptual experience. Under this account, strong imagers generate richer predictive models that more readily engage higher-order visual representations, while weak imagers rely more heavily on bottom-up input. The Ganzflicker paradigm—which provides rhythmic stimulation with minimal structured input—may be particularly sensitive to these differences: when bottom-up information is impoverished, individual variation in top-down generative capacity becomes visible in the content of conscious experience.

Beyond Ganzflicker, these findings may have broader implications for understanding human cognition. Individual differences in imagery are not merely subjective curiosities but have been shown to influence a wide range of cognitive processes, from shaping our emotions (Wicken et al., 2021) and moral judgments (Amit & Greene, 2012) to affecting how we adopt scientific theories (Sulik et al., 2025). By incorporating phenomenological variability into our models, we move toward a more complete and inclusive science of the mind. Further, this work demonstrates that computational analysis of free-text reports can reveal the hidden structure of complex mental phenomena, offering a scalable approach for studying individual differences in cognitive processes that are otherwise difficult to access (Feuerriegel et al., 2025). By treating language as a window into the mind's eye, we can begin to capture how people with different imagery phenotypes internally construct their visual world.

## Materials availability

Ganzflicker experience and questionnaires can be found at:
https://forms.gle/tdKRKhva3uqC68tS9

## Data availability

The human dataset of the Ganzflicker study is from (Reeder, 2022), and is accessible at: https://osf.io/6dvh9/, and the cleaned dataset we used in this paper is available at:
https://github.com/anachkhaidze/ganzflickerNLPalooza_paper/tree/main/data
The Lancaster Norms is from (Lynott et al. (2020), and the data can be found at:
https://github.com/anachkhaidze/ganzflickerNLPalooza_paper/tree/main/data

## Code availability

Analysis scripts (in *R* and *Python*) for all components of the study, including topic modeling and Lancaster Sensorimotor Norms analysis, are available at:
https://github.com/anachkhaidze/ganzflickerNLPalooza_paper

The full pipeline is procedurally reproducible: all scripts, configuration files, and random seeds are provided to enable replication. Minor numerical variation may occur due to stochasticity in embedding models; however, the overall topic structure and model-derived similarity patterns remain stable across runs.

## Author contributions

A.C., S.C., A.K. and R.R.R. conceived the project. A.C., S.C. and C.G. designed the analyses. A.C. and C.G. conducted the analyses. S.C. supervised the project. A.C. and S.C. wrote the original draft. All others contributed to revising the manuscript and provided critical feedback.

## Acknowledgements

The authors thank Sean Trott for helpful suggestions regarding statistical analysis, and the Brain & Cognition Lab and the Cognitive Neuroscience Lab at UC San Diego for valuable feedback and discussion.

## Funding

This research was supported by an Innovative Research Grant from the Kavli Institute for Brain and Mind at UC San Diego awarded to A.C., S.C., and A.K.

## Competing interests

The authors declare no competing interests.

# References

Albers, A. M., Kok, P., Toni, I., Dijkerman, H. C., & de Lange, F. P. (2013). Shared representations for working memory and mental imagery in early visual cortex. *Current Biology: CB*, *23*(15), 1427–1431. https://doi.org/10.1016/j.cub.2013.05.065

Allefeld, C., Pütz, P., Kastner, K., & Wackermann, J. (2011). Flicker-light induced visual phenomena: Frequency dependence and specificity of whole percepts and percept features. *Consciousness and Cognition*, *20*(4), 1344–1362. https://doi.org/10.1016/j.concog.2010.10.026

Amaya, I. A., Nierhaus, T., & Schmidt, T. T. (2025). Thalamocortical interactions reflecting the intensity of flicker light–induced visual hallucinatory phenomena. *Network Neuroscience*, *9*(1), 1–17. https://doi.org/10.1162/netn_a_00417

Amaya, I. A., Behrens, N., Schwartzman, D. J., Hewitt, T., & Schmidt, T. T. (2023). Effect of frequency and rhythmicity on flicker light-induced hallucinatory phenomena. *PLOS ONE*, *18*(4), e0284271. https://doi.org/10.1371/journal.pone.0284271

Amit, E., & Greene, J. D. (2012). You See, the Ends Don't Justify the Means: Visual Imagery and Moral Judgment. *Psychological Science*, *23*(8), 861–868. https://doi.org/10.1177/0956797611434965

Aydın, Ç. (2020). Gender differences in visual imagery: Object and spatial imagery. *Dokuz Eylül Üniversitesi Sosyal Bilimler Enstitüsü Dergisi*, *22*(3), 1045–1064. https://doi.org/10.16953/deusosbil.531463

Bainbridge, W. A., Pounder, Z., Eardley, A. F., & Baker, C. I. (2021). Quantifying aphantasia through drawing: Those without visual imagery show deficits in object but not spatial memory. *Cortex*, *135*, 159–172. https://doi.org/10.1016/j.cortex.2020.11.014

Banks, B., & Connell, L. (2022). Multi-dimensional sensorimotor grounding of concrete and abstract categories. *Philosophical Transactions of the Royal Society B: Biological Sciences*, *378*(1870), 20210366. https://doi.org/10.1098/rstb.2021.0366

Barrus, T. (2025). *Barrust/pyspellchecker* [Python]. https://github.com/barrust/pyspellchecker (Original work published 2018)

Bartossek, M. T., Kemmerer, J., & Schmidt, T. T. (2021). Altered states phenomena induced by visual flicker light stimulation. *PLOS ONE*, *16*(7), e0253779. https://doi.org/10.1371/journal.pone.0253779

Beauté, R., Schwartzman, D. J., Dumas, G., Crook, J., Macpherson, F., Barrett, A. B., & Seth, A. K. (2025). *Mapping of Subjective Accounts into Interpreted Clusters (MOSAIC): Topic Modelling and LLM applied to Stroboscopic Phenomenology* (arXiv:2502.18318). arXiv. https://doi.org/10.48550/arXiv.2502.18318

Bergmann, J., Genç, E., Kohler, A., Singer, W., & Pearson, J. (2016). Smaller Primary Visual Cortex Is Associated with Stronger, but Less Precise Mental Imagery. *Cerebral Cortex*, *26*(9), 3838–3850. https://doi.org/10.1093/cercor/bhv186

Berkovich-Ohana, A., Dor-Ziderman, Y., Glicksohn, J., & Goldstein, A. (2013). Alterations in the sense of time, space, and body in the mindfulness-trained brain: A neurophenomenologically-guided MEG study. *Frontiers in Psychology*, *4*. https://doi.org/10.3389/fpsyg.2013.00912

Bird, C. M., Bisby, J. A., & Burgess, N. (2012). The hippocampus and spatial constraints on mental imagery. *Frontiers in Human Neuroscience*, *6*. https://doi.org/10.3389/fnhum.2012.00142

Bird, S., Klein, E., & Loper, E. (2009). *Natural Language Processing with Python: Analyzing Text with the Natural Language Toolkit*. O'Reilly Media, Inc.

Blei, D. M., Ng, A. Y., & Jordan, M. I. (2003). Latent dirichlet allocation. *J. Mach. Learn. Res.*, *3*(0), 993–1022.

Bracci, S., Ietswaart, M., Peelen, M. V., & Cavina-Pratesi, C. (2010). Dissociable neural responses to hands and non-hand body parts in human left extrastriate visual cortex. *Journal of Neurophysiology*, *103*(6), 3389–3397. https://doi.org/10.1152/jn.00215.2010

Bressloff, P. C., Cowan, J. D., Golubitsky, M., Thomas, P. J., & Wiener, M. C. (2001). Geometric visual hallucinations, Euclidean symmetry and the functional architecture of striate cortex. *Philosophical Transactions of the Royal Society of London. Series B, Biological Sciences*, *356*(1407), 299–330. https://doi.org/10.1098/rstb.2000.0769

Bressloff, P. C., Cowan, J. D., Golubitsky, M., Thomas, P. J., & Wiener, M. C. (2002). What Geometric Visual Hallucinations Tell Us about the Visual Cortex. *Neural Computation*, *14*(3), 473–491. https://doi.org/10.1162/089976602317250861

Burke, W. (2002). The neural basis of Charles Bonnet hallucinations: A hypothesis. *Journal of Neurology, Neurosurgery, and Psychiatry*, *73*(5), 535–541. https://doi.org/10.1136/jnnp.73.5.535

Cabbai, G., Racey, C., Simner, J., Dance, C., Ward, J., & Forster, S. (2024). Sensory representations in primary visual cortex are not sufficient for subjective imagery. *Current Biology*, *34*(21), 5073-5082.e5. https://doi.org/10.1016/j.cub.2024.09.062

Campello, R. J. G. B., Moulavi, D., & Sander, J. (2013). Density-Based Clustering Based on Hierarchical Density Estimates. In J. Pei, V. S. Tseng, L. Cao, H. Motoda, & G. Xu (Eds.), *Advances in Knowledge Discovery and Data Mining* (pp. 160–172). Springer. https://doi.org/10.1007/978-3-642-37456-2_14

Chang, S., Zhang, X., Cao, Y., Pearson, J., & Meng, M. (2025). Imageless imagery in aphantasia revealed by early visual cortex decoding. *Current Biology*, *35*(3), 591-599.e4. https://doi.org/10.1016/j.cub.2024.12.012

Chkhaidze, A., Coulson, S., & Kiyonaga, A. (2023). Individual Differences in Preferred Thought Formats Predict Features of Narrative Recall. *Proceedings of the Annual Meeting of the Cognitive Science Society*, *45*(45). https://escholarship.org/uc/item/0jm2d2rm

Christian, B. M., Miles, L. K., Parkinson, C., & Macrae, C. N. (2013). Visual perspective and the characteristics of mind wandering. *Frontiers in Psychology*, *4*, 699. https://doi.org/10.3389/fpsyg.2013.00699

Connell, L., & Lynott, D. (2012). Strength of perceptual experience predicts word processing performance better than concreteness or imageability. *Cognition*, *125*(3), 452–465. https://doi.org/10.1016/j.cognition.2012.07.010

Cui, X., Jeter, C. B., Yang, D., Montague, P. R., & Eagleman, D. M. (2007). Vividness of mental imagery: Individual variability can be measured objectively. *Vision Research*, *47*(4), 474–478. https://doi.org/10.1016/j.visres.2006.11.013

Dance, C. J., Ward, J., & Simner, J. (2021). What is the Link Between Mental Imagery and Sensory Sensitivity? Insights from Aphantasia. *Perception*, *50*(9), 757–782. https://doi.org/10.1177/03010066211042186

Dijkstra, N., Bosch, S. E., & van Gerven, M. A. J. (2019). Shared Neural Mechanisms of Visual Perception and Imagery. *Trends in Cognitive Sciences*, *23*(5), 423–434. https://doi.org/10.1016/j.tics.2019.02.004

Epstein, R., & Kanwisher, N. (1998). A cortical representation of the local visual environment. *Nature*, *392*(6676), 598–601. https://doi.org/10.1038/33402

Ermentrout, G. B., & Cowan, J. D. (1979). A mathematical theory of visual hallucination patterns. *Biological Cybernetics*, *34*(3), 137–150. https://doi.org/10.1007/BF00336965

Feuerriegel, S., Maarouf, A., Bär, D., Geissler, D., Schweisthal, J., Pröllochs, N., Robertson, C. E., Rathje, S., Hartmann, J., Mohammad, S. M., Netzer, O., Siegel, A. A., Plank, B., & Van Bavel, J. J. (2025). Using natural language processing to analyse text data in behavioural science. *Nature Reviews Psychology*, *4*(2), 96–111. https://doi.org/10.1038/s44159-024-00392-z

ffytche, D. H. (2008). The hodology of hallucinations. *Cortex; a Journal Devoted to the Study of the Nervous System and Behavior*, *44*(8), 1067–1083. https://doi.org/10.1016/j.cortex.2008.04.005

Fort, L. D., Costines, C., Wittmann, M., Demertzi, A., & Schmidt, T. T. (2025). Classification schemes of altered states of consciousness. *Neuroscience & Biobehavioral Reviews*, *175*, 106178. https://doi.org/10.1016/j.neubiorev.2025.106178

Funder, D. C., & Ozer, D. J. (2019). Evaluating Effect Size in Psychological Research: Sense and Nonsense. *Advances in Methods and Practices in Psychological Science*, *2*(2), 156–168. https://doi.org/10.1177/2515245919847202

Grootendorst, M. (2022). *BERTopic: Neural topic modeling with a class-based TF-IDF procedure*. arXiv.Org. https://arxiv.org/abs/2203.05794v1

Gulyás, E., Gombos, F., Sütöri, S., Lovas, A., Ziman, G., & Kovács, I. (2022). Visual imagery vividness declines across the lifespan. *Cortex; a Journal Devoted to the Study of the Nervous System and Behavior*, *154*, 365–374. https://doi.org/10.1016/j.cortex.2022.06.011

Hewitt, T., Amaya, I., Beauté, R., Seth, A. K., Schmidt, T. T., & Schwartzman, D. J. (2025). Stroboscopically induced visual hallucinations: Historical, phenomenological, and neurobiological perspectives. *Neuroscience of Consciousness*, *2025*(1), niaf020. https://doi.org/10.1093/nc/niaf020

Hwang, K., Bertolero, M. A., Liu, W. B., & D'Esposito, M. (2017). The Human Thalamus Is an Integrative Hub for Functional Brain Networks. *Journal of Neuroscience*, *37*(23), 5594–5607. https://doi.org/10.1523/JNEUROSCI.0067-17.2017

Kanwisher, N., McDermott, J., & Chun, M. M. (1997). The Fusiform Face Area: A Module in Human Extrastriate Cortex Specialized for Face Perception. *Journal of Neuroscience*, *17*(11), 4302–4311. https://doi.org/10.1523/JNEUROSCI.17-11-04302.1997

Keogh, R., & Pearson, J. (2014). The sensory strength of voluntary visual imagery predicts visual working memory capacity. *Journal of Vision*, *14*(12), 7. https://doi.org/10.1167/14.12.7

Kiss, T., & Strunk, J. (2006). Unsupervised Multilingual Sentence Boundary Detection. *Computational Linguistics*, *32*(4), 485–525. https://doi.org/10.1162/coli.2006.32.4.485

Klüver, H. (1926). Mescal Visions and Eidetic Vision. *The American Journal of Psychology*, *37*(4), 502–515. https://doi.org/10.2307/1414910

Königsmark, V. T., Bergmann, J., & Reeder, R. R. (2021). The Ganzflicker experience: High probability of seeing vivid and complex pseudo-hallucinations with imagery but not aphantasia. *Cortex*, *141*, 522–534. https://doi.org/10.1016/j.cortex.2021.05.007

Kosslyn, S. M., Brunn, J., Cave, K. R., & Wallach, R. W. (1984). Individual differences in mental imagery ability: A computational analysis. *Cognition*, *18*(1), 195–243. https://doi.org/10.1016/0010-0277(84)90025-8

Kübel, S. L., Fiedler, H., & Wittmann, M. (2021). Red visual stimulation in the Ganzfeld leads to a relative overestimation of duration compared to green. *PsyCh Journal*, *10*(1), 5–19. https://doi.org/10.1002/pchj.395

Lacruz, R. S., Stringer, C. B., Kimbel, W. H., Wood, B., Harvati, K., O'Higgins, P., Bromage, T. G., & Arsuaga, J.-L. (2019). The evolutionary history of the human face. *Nature Ecology & Evolution*, *3*(5), 726–736. https://doi.org/10.1038/s41559-019-0865-7

Lee, S.-H., Kravitz, D. J., & Baker, C. I. (2012). Disentangling visual imagery and perception of real-world objects. *NeuroImage*, *59*(4), 4064–4073. https://doi.org/10.1016/j.neuroimage.2011.10.055

Liu, J., Zhan, M., Hajhajate, D., Spagna, A., Dehaene, S., Cohen, L., & Bartolomeo, P. (2025). Visual mental imagery in typical imagers and in aphantasia: A millimeter-scale 7-T fMRI study. *Cortex*, *185*, 113–132. https://doi.org/10.1016/j.cortex.2025.01.013

Lupyan, G., Uchiyama, R., Thompson, B., & Casasanto, D. (2023). Hidden Differences in Phenomenal Experience. *Cognitive Science*, *47*(1), e13239. https://doi.org/10.1111/cogs.13239

Lynott, D., Connell, L., Brysbaert, M., Brand, J., & Carney, J. (2020). The Lancaster Sensorimotor Norms: Multidimensional measures of perceptual and action strength for 40,000 English words. *Behavior Research Methods*, *52*(3), 1271–1291. https://doi.org/10.3758/s13428-019-01316-z

Marks, D. F. (1973). Visual Imagery Differences in the Recall of Pictures. *British Journal of Psychology*, *64*(1), 17–24. https://doi.org/10.1111/j.2044-8295.1973.tb01322.x

McInnes, L., Healy, J., & Astels, S. (2017). hdbscan: Hierarchical density based clustering. *The Journal of Open Source Software*, *2*, 205. https://doi.org/10.21105/joss.00205

McInnes, L., Healy, J., & Melville, J. (2018). *UMAP: Uniform Manifold Approximation and Projection for Dimension Reduction* (arXiv:1802.03426). arXiv. https://doi.org/10.48550/arXiv.1802.03426

Mechelli, A., Price, C. J., Friston, K. J., & Ishai, A. (2004). Where Bottom-up Meets Top-down: Neuronal Interactions during Perception and Imagery. *Cerebral Cortex*, *14*(11), 1256–1265. https://doi.org/10.1093/cercor/bhh087

Merabet, L. B., Maguire, D., Warde, A., Alterescu, K., Stickgold, R., & Pascual-Leone, A. (2004). Visual Hallucinations During Prolonged Blindfolding in Sighted Subjects. *Journal of Neuro-Ophthalmology*, *24*(2), 109.

Metzger, W. (1929). Optische Untersuchungen am Ganzfeld. II. Zur Phänomenologie des homogenen Ganzfelds. *Psychologische Forschung*, *13*, 6–29. https://doi.org/10.1007/BF00406757

Milton, F., Fulford, J., Dance, C., Gaddum, J., Heuerman-Williamson, B., Jones, K., Knight, K. F., MacKisack, M., Winlove, C., & Zeman, A. (2021). Behavioral and Neural Signatures of Visual Imagery Vividness Extremes: Aphantasia versus Hyperphantasia. *Cerebral Cortex Communications*, *2*(2), tgab035. https://doi.org/10.1093/texcom/tgab035

Muraki, E. J., Speed, L. J., & Pexman, P. M. (2023). Insights into embodied cognition and mental imagery from aphantasia. *Nature Reviews Psychology*, *2*(10), 591–605. https://doi.org/10.1038/s44159-023-00221-9

Nichols, T. E., & Holmes, A. P. (2002). Nonparametric permutation tests for functional neuroimaging: A primer with examples. *Human Brain Mapping*, *15*(1), 1–25. https://doi.org/10.1002/hbm.1058

O'Callaghan, D., Greene, D., Carthy, J., & Cunningham, P. (2015). An analysis of the coherence of descriptors in topic modeling. *Expert Systems with Applications*, *42*(13), 5645–5657. https://doi.org/10.1016/j.eswa.2015.02.055

OpenAI. (2024). GPT-4o-mini [Large language model]. https://platform.openai.com/

Pearson, J. (2019). The human imagination: The cognitive neuroscience of visual mental imagery. *Nature Reviews Neuroscience*, *20*(10), 624–634. https://doi.org/10.1038/s41583-019-0202-9

Pedregosa, F., Varoquaux, G., Gramfort, A., Michel, V., Thirion, B., Grisel, O., Blondel, M., Prettenhofer, P., Weiss, R., Dubourg, V., Vanderplas, J., Passos, A., Cournapeau, D., Brucher, M., Perrot, M., & Duchesnay, E. (2011). Scikit-learn: Machine Learning in Python. *Journal of Machine Learning Research*. https://inria.hal.science/hal-00650905

Phillips, J. W., Schulmann, A., Hara, E., Winnubst, J., Liu, C., Valakh, V., Wang, L., Shields, B. C., Korff, W., Chandrashekar, J., Lemire, A. L., Mensh, B., Dudman, J. T., Nelson, S. B., & Hantman, A. W. (2019). A repeated molecular architecture across thalamic pathways. *Nature Neuroscience*, *22*(11), 1925–1935. https://doi.org/10.1038/s41593-019-0483-3

Pistolas, E., & Wagemans, J. (2025). And then there was light in the ganzfeld: Clarifying the methods, experiences, and modulating factors of hallucinations and decays. *Neuroscience of Consciousness*, *2025*(1). https://doi.org/10.1093/nc/niaf021

Purkinje, J. E. (1819). *Beiträge Zur Kenntniss Des Sehens in Subjectiver Hinsicht*. Johann Gottfried Calve. http://archive.org/details/beitrgezurkenn00purk

Pütz, P., Braeunig, M., & Wackermann, J. (2006). EEG correlates of multimodal ganzfeld induced hallucinatory imagery. *International Journal of Psychophysiology*, *61*(2), 167–178. https://doi.org/10.1016/j.ijpsycho.2005.09.002

Python Software Foundation. (2023). Python programming language. https://www.python.org/

R Core Team. (2023). R: A language and environment for statistical computing. R Foundation for Statistical Computing.

Reeder, R. (2021, June 22). *Pseudo-hallucinations: Why some people see more vivid mental images than others – test yourself here*. The Conversation. http://theconversation.com/pseudo-hallucinations-why-some-people-see-more-vivid-mental-images-than-others-test-yourself-here-163025

Reeder, R. R. (2022). Ganzflicker Reveals the Complex Relationship Between Visual Mental Imagery and Pseudo-Hallucinatory Experiences: A Replication and Expansion. *Collabra: Psychology*, *8*(1), 36318. https://doi.org/10.1525/collabra.36318

Reeder, R. R. (2024). Current and future directions for research on hallucinations and delusions. *Scientific Reports*, *14*(1), 8328. https://doi.org/10.1038/s41598-024-57472-6

Reeder, R. R., Sala, G., & van Leeuwen, T. M. (2024). A novel model of divergent predictive perception. *Neuroscience of Consciousness*, *2024*(1), niae006. https://doi.org/10.1093/nc/niae006

Reimers, N., & Gurevych, I. (2019). *Sentence-BERT: Sentence Embeddings using Siamese BERT-Networks* (arXiv:1908.10084). arXiv. https://doi.org/10.48550/arXiv.1908.10084

Richardson, J. T. E. (1995). Gender differences in the Vividness of Visual Imagery Questionnaire: A meta-analysis. *Journal of Mental Imagery*, *19*(3–4), 177–187.

Röder, M., Both, A., & Hinneburg, A. (2015). Exploring the Space of Topic Coherence Measures. *Proceedings of the Eighth ACM International Conference on Web Search and Data Mining*, 399–408. https://doi.org/10.1145/2684822.2685324

Rule, M., Stoffregen, M., & Ermentrout, B. (2011). A Model for the Origin and Properties of Flicker-Induced Geometric Phosphenes. *PLOS Computational Biology*, *7*(9), e1002158. https://doi.org/10.1371/journal.pcbi.1002158

Saito, T., & Rehmsmeier, M. (2015). The Precision-Recall Plot Is More Informative than the ROC Plot When Evaluating Binary Classifiers on Imbalanced Datasets. *PLOS ONE*, *10*(3), e0118432. https://doi.org/10.1371/journal.pone.0118432

Schmidt, T. T., Jagannathan, N., Ljubljanac, M., Xavier, A., & Nierhaus, T. (2020). The multimodal Ganzfeld-induced altered state of consciousness induces decreased thalamo-cortical coupling. *Scientific Reports*, *10*(1), 18686. https://doi.org/10.1038/s41598-020-75019-3

Schmidt, T. T., & Prein, J. C. (2019). The Ganzfeld experience—A stably inducible altered state of consciousness: Effects of different auditory homogenizations. *PsyCh Journal*, *8*(1), 66–81. https://doi.org/10.1002/pchj.262

Schwartzman, D. J., Schartner, M., Ador, B. B., Simonelli, F., Chang, A. Y.-C., & Seth, A. K. (2019). *Increased spontaneous EEG signal diversity during stroboscopically-induced altered states of consciousness* (p. 511766). bioRxiv. https://doi.org/10.1101/511766

Shenyan, O., Lisi, M., Greenwood, J. A., Skipper, J. I., & Dekker, T. M. (2024). Visual hallucinations induced by Ganzflicker and Ganzfeld differ in frequency, complexity, and content. *Scientific Reports*, *14*(1), 2353. https://doi.org/10.1038/s41598-024-52372-1

Shine, J. M. (2021). The thalamus integrates the macrosystems of the brain to facilitate complex, adaptive brain network dynamics. *Progress in Neurobiology*, *199*, 101951. https://doi.org/10.1016/j.pneurobio.2020.101951

Silvanto, J. (2025). Interoception, insula, and agency: A predictive coding account of aphantasia. *Frontiers in Psychology*, *16*. https://doi.org/10.3389/fpsyg.2025.1564251

Silvanto, J., & Nagai, Y. (2025). How Interoception and the Insula Shape Mental Imagery and Aphantasia. *Brain Topography*, *38*(2), 27. https://doi.org/10.1007/s10548-025-01101-6

Spagna, A., Hajhajate, D., Liu, J., & Bartolomeo, P. (2021). Visual mental imagery engages the left fusiform gyrus, but not the early visual cortex: A meta-analysis of neuroimaging evidence. *Neuroscience & Biobehavioral Reviews*, *122*, 201–217. https://doi.org/10.1016/j.neubiorev.2020.12.029

Stoliker, D., Preller, K. H., Novelli, L., Anticevic, A., Egan, G. F., Vollenweider, F. X., & Razi, A. (2025). Neural mechanisms of psychedelic visual imagery. *Molecular Psychiatry*, *30*(4), 1259–1266. https://doi.org/10.1038/s41380-024-02632-3

Sulik, J., Rim, N., Pontikes, E., Evans, J., & Lupyan, G. (2025). Differences in psychologists' cognitive traits are associated with scientific divides. *Nature Human Behaviour*, 1–15. https://doi.org/10.1038/s41562-025-02153-1

Suzuki, M., & Larkum, M. E. (2020). General Anesthesia Decouples Cortical Pyramidal Neurons. *Cell*, *180*(4), 666-676.e13. https://doi.org/10.1016/j.cell.2020.01.024

Tibshirani, R. (1996). Regression Shrinkage and Selection Via the Lasso. *Journal of the Royal Statistical Society: Series B (Methodological)*, *58*(1), 267–288. https://doi.org/10.1111/j.2517-6161.1996.tb02080.x

Tingley, D., Yamamoto, T., Hirose, K., Keele, L., & Imai, K. (2014). mediation: R Package for Causal Mediation Analysis. *Journal of Statistical Software*, *59*, 1–38. https://doi.org/10.18637/jss.v059.i05

Vinaya, H., Trott, S., Pecher, D., Zeelenberg, R., & Coulson, S. (2025). Words and Worlds Both: Dynamic Effects of Distributional and Sensorimotor Information in Semantic Processing. *Open Mind*, *9*, 2149–2174. https://doi.org/10.1162/OPMI.a.316

Wackermann, J., Pütz, P., & Allefeld, C. (2008). Ganzfeld-induced hallucinatory experience, its phenomenology and cerebral electrophysiology. *Cortex*, *44*(10), 1364–1378. https://doi.org/10.1016/j.cortex.2007.05.003

Wicken, M., Keogh, R., & Pearson, J. (2021). The critical role of mental imagery in human emotion: Insights from fear-based imagery and aphantasia. *Proceedings. Biological Sciences*, *288*(1946), 20210267. https://doi.org/10.1098/rspb.2021.0267

Zeman, A. (2024). Aphantasia and hyperphantasia: Exploring imagery vividness extremes. *Trends in Cognitive Sciences*, *28*(5), 467–480. https://doi.org/10.1016/j.tics.2024.02.007

Zeman, A., Milton, F., Della Sala, S., Dewar, M., Frayling, T., Gaddum, J., Hattersley, A., Heuerman-Williamson, B., Jones, K., MacKisack, M., & Winlove, C. (2020). Phantasia–The psychological significance of lifelong visual imagery vividness extremes. *Cortex*, *130*, 426–440. https://doi.org/10.1016/j.cortex.2020.04.003

## Supplementary material

### 1 Example hallucination descriptions

*"Swirls, a faded, ghost-like tulip that moved in circular ways, balls of light. Later, there were straight, line like fragments, black ones around corners. The red combined with sunset like colors to yellow on edges … I was briefly reminded of a total solar eclipse I saw a few years back, not sure why … Screen expanding and contracting, lines changing into small squares moving into center of screen … I also then tried to create something to see, that did not work, all was random"*

*"Saw old stone buildings, first a small square room spinning and then i was outside of the room and the building was spinning, later i started seeing like the outside of the building (it was like it was a castle?) there were some stone columns and all was blue-ish stone, someone was standing looking at me but it was just a shadow, i was flying above it so the person was small and I couldn't see anything."*

*"I saw purple dots that were circling around like stars in a galaxy, also with spiral arms etc. They were going quite fast, I wanted them to slow down (to imitate a galaxy) but they didn't. Then in the next few seconds the purple dots would form other shapes like waves or pillars that were turning as well."*

*"There was this bird, almost like a Native American symbol, rising up from the SE corner that was blue and gold, looking down on a man in armor with his back to me holding a sword. Another was a space station—futuristic and circular with ships flying in and out like a port, then eventually meteors crashing into it. I also kept seeing various images of knights and soldiers—war and fighting."*

*"It started with a sort of a fractal, coming out of a white circle in the middle, with eight grey branches coming outwards, which then split into 8 more, etc."*

*"I felt like my eyes were closed, but I could still see all this,the triangles violently shifting, breaking into a circle of white, surrounded by a cyan halo. It was around then the white noise stopped, and the circle turned into a heart, a black outline around a red and white circle."*

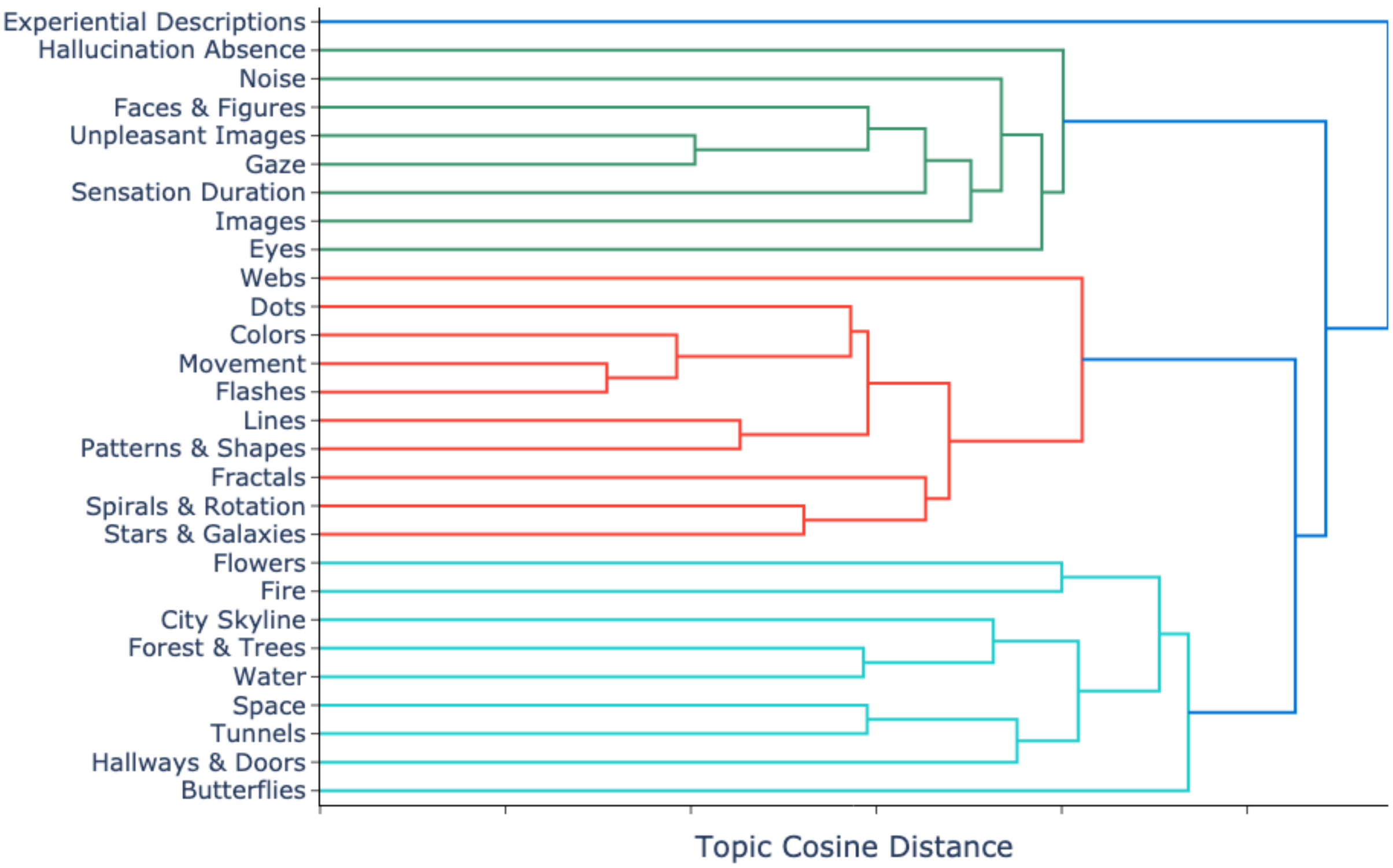

**Figure S1. Dendrogram illustrating hierarchical relationships among topics**. The dendrogram was constructed using cosine distances between topic embeddings, revealing the hierarchical semantic structure that emerges from the topic modeling process. The clustering revealed three emergent experiential clusters: social and metacognitive (green), simple visuals (red), naturalistic scenes (cyan).

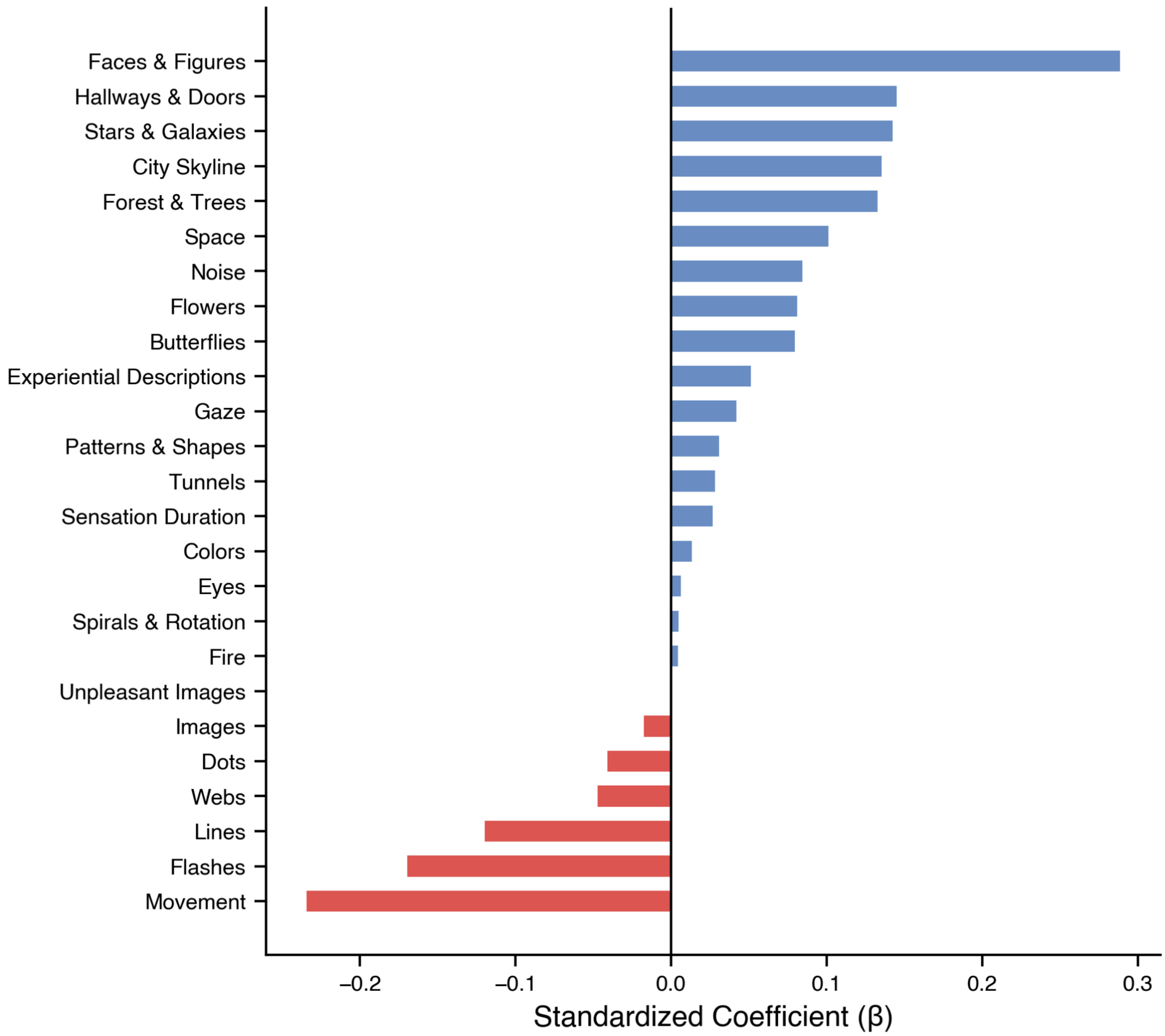

**Figure S2. Lasso regression coefficients for continuous visual vividness prediction**. Forest plot showing standardized Lasso regression coefficients (β) for continuous vividness prediction. Only non-zero coefficients after cross-validated L1 regularization are displayed (n = 25 of 28 topics). Positive coefficients (blue bars) indicate topics associated with higher self-reported vividness; negative coefficients (red bars) indicate topics associated with lower vividness. Complex visual scenes and dynamic imagery were most predictive of higher vividness scores, while basic visual elements and perceptual distortions were associated with lower scores. The model achieved $R^2 = .05$ (MSE = 8.63) with optimal $\alpha = 0.03$.

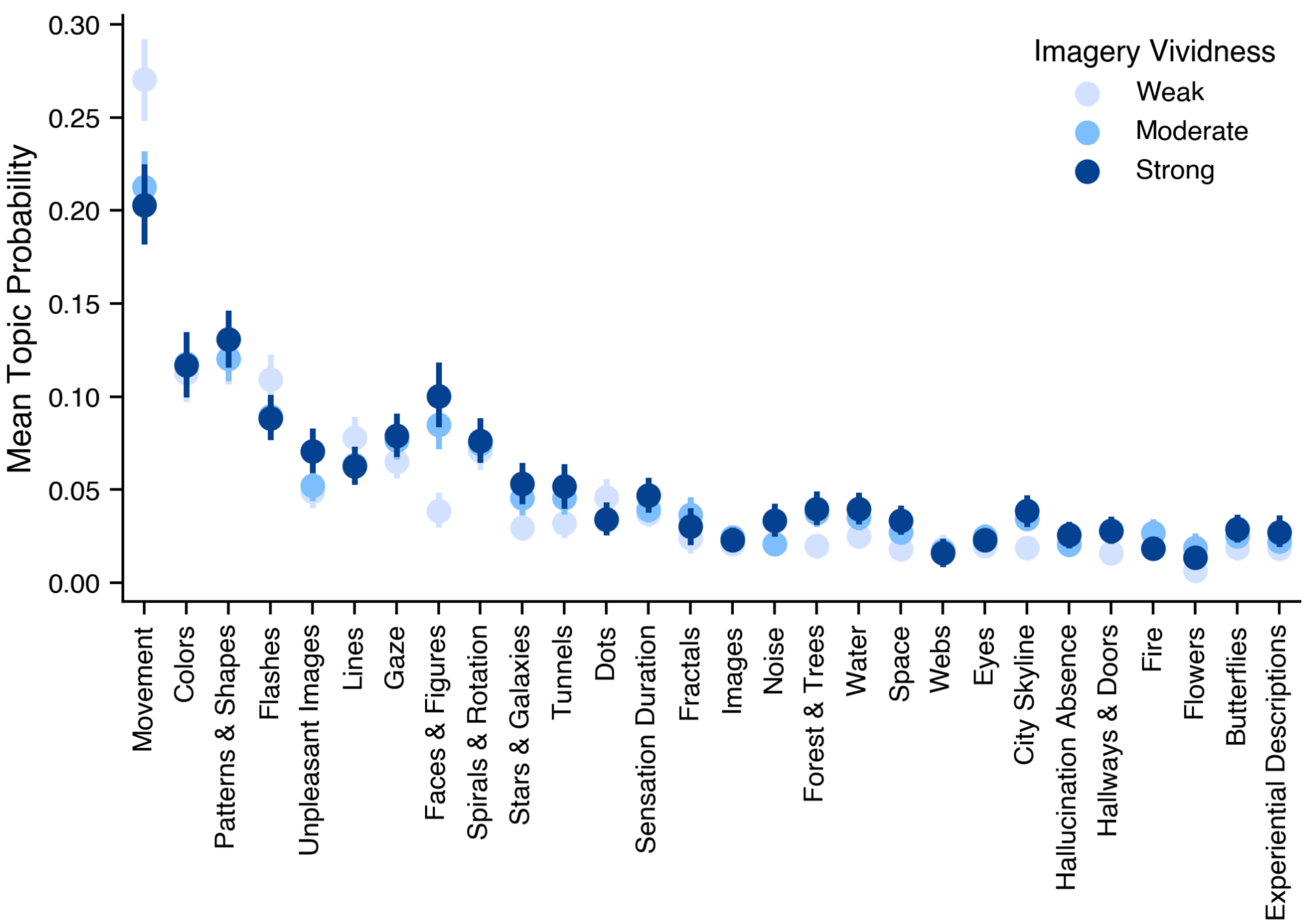

**Figure S3. Imagery group-level topic expression**. Mean topic probabilities (±95% CI) by imagery vividness group. Strong imagers were more likely to describe complex content (e.g., *faces, natural and urban scenes*) compared to weak imagers, while weak imagers described simpler visual phenomena (e.g., *visual flashes, lines*), moderate imagers' descriptions showed less coherence.

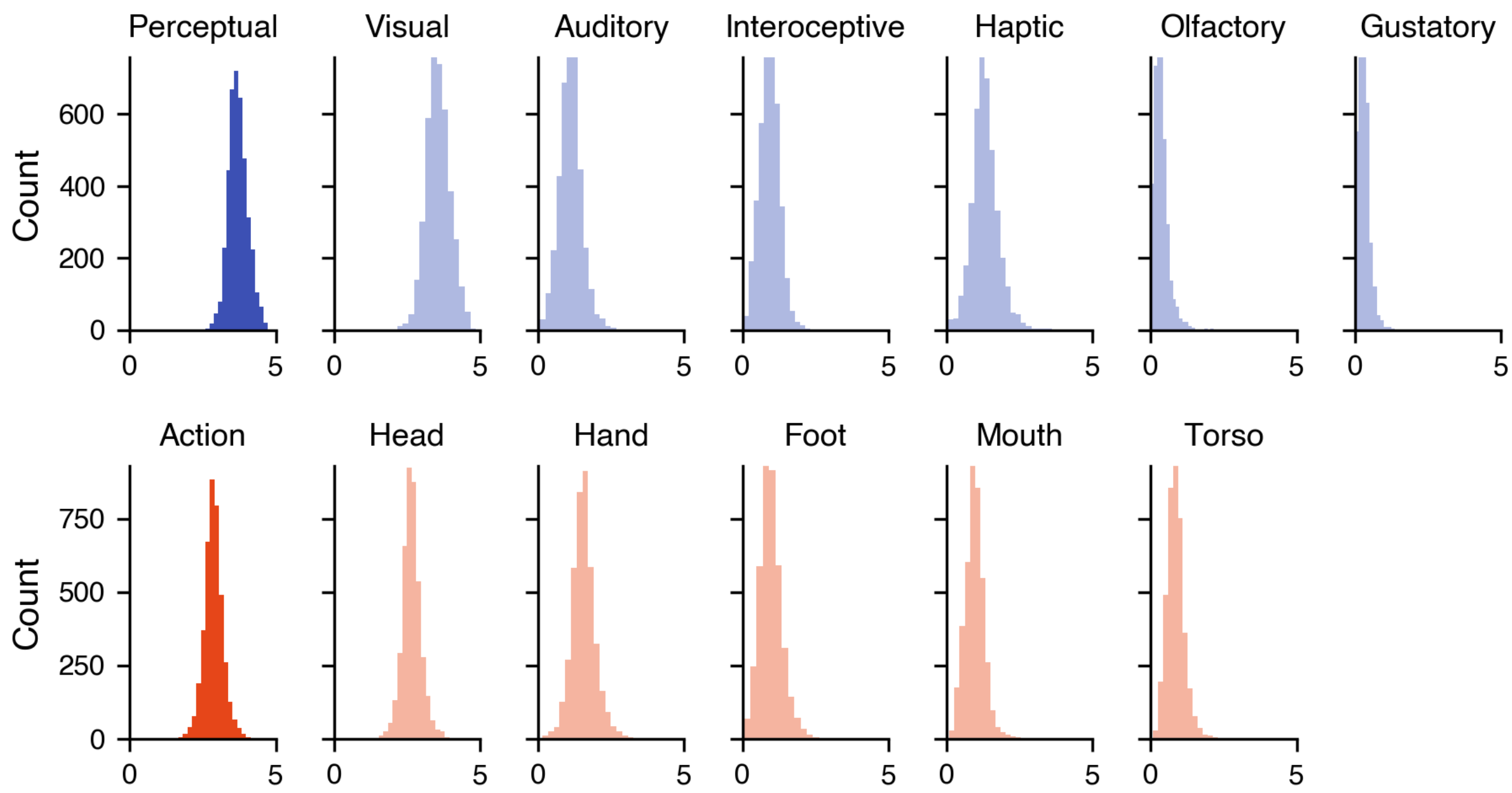

**Figure S4. Distributions of Lancaster Sensorimotor Norm dimensions across hallucination descriptions**. Histograms show the distribution of values for 13 sensorimotor dimensions drawn from the Lancaster Sensorimotor Norms, averaged across words within each hallucination description. Dimensions are grouped by type: sensory (top row) and motor dimensions (bottom row). Solid-colored histograms represent composite measures of perceptual strength (dark blue) and action strength (dark red); lighter shades indicate specific subdimensions. These plots reflect the range and density of embodied and perceptual content present in the dataset. As shown, perceptual and motor variables vary in overall magnitude and dispersion, with visual terms dominating perceptual content, and head- and hand-related content dominating motor content of the hallucination descriptions.

**Table S1. Descriptive statistics for Lancaster Sensorimotor Norm dimensions.**
Mean, range, and variance for each sensorimotor dimension computed across all hallucination descriptions (n = 4,057). Higher scores indicate stronger association with the corresponding modality or body part. Composite scores (perceptual and motor strength) reflect the scores of the dominant perceptual and motor modalities, respectively.

| Dimension | Mean | Min | Max | Range | Variance |
|---|---|---|---|---|---|
| **Perceptual** | | | | | |
| Visual Strength | 3.57 | 1.00 | 4.87 | 3.87 | 0.16 |
| Haptic Strength | 1.32 | 0.05 | 3.62 | 3.57 | 0.19 |
| Auditory Strength | 1.11 | 0.06 | 3.79 | 3.73 | 0.14 |
| Interoceptive Strength | 0.93 | 0.05 | 3.53 | 3.48 | 0.12 |
| Olfactory Strength | 0.36 | 0.00 | 2.16 | 2.16 | 0.06 |
| Gustatory Strength | 0.30 | 0.00 | 2.48 | 2.48 | 0.04 |
| **Motor** | | | | | |
| Head Strength | 2.66 | 1.53 | 4.76 | 3.23 | 0.10 |
| Hand Strength | 1.56 | 0.14 | 4.05 | 3.91 | 0.17 |
| Foot Strength | 0.97 | 0.05 | 4.36 | 4.31 | 0.15 |
| Mouth Strength | 0.96 | 0.10 | 3.60 | 3.5 | 0.11 |
| Torso Strength | 0.85 | 0.08 | 3.48 | 3.4 | 0.09 |
| **Composite Measures** | | | | | |
| Perceptual Strength | 3.71 | 2.00 | 4.87 | 2.87 | 0.11 |
| Motor Strength | 2.87 | 1.66 | 4.76 | 3.1 | 0.11 |

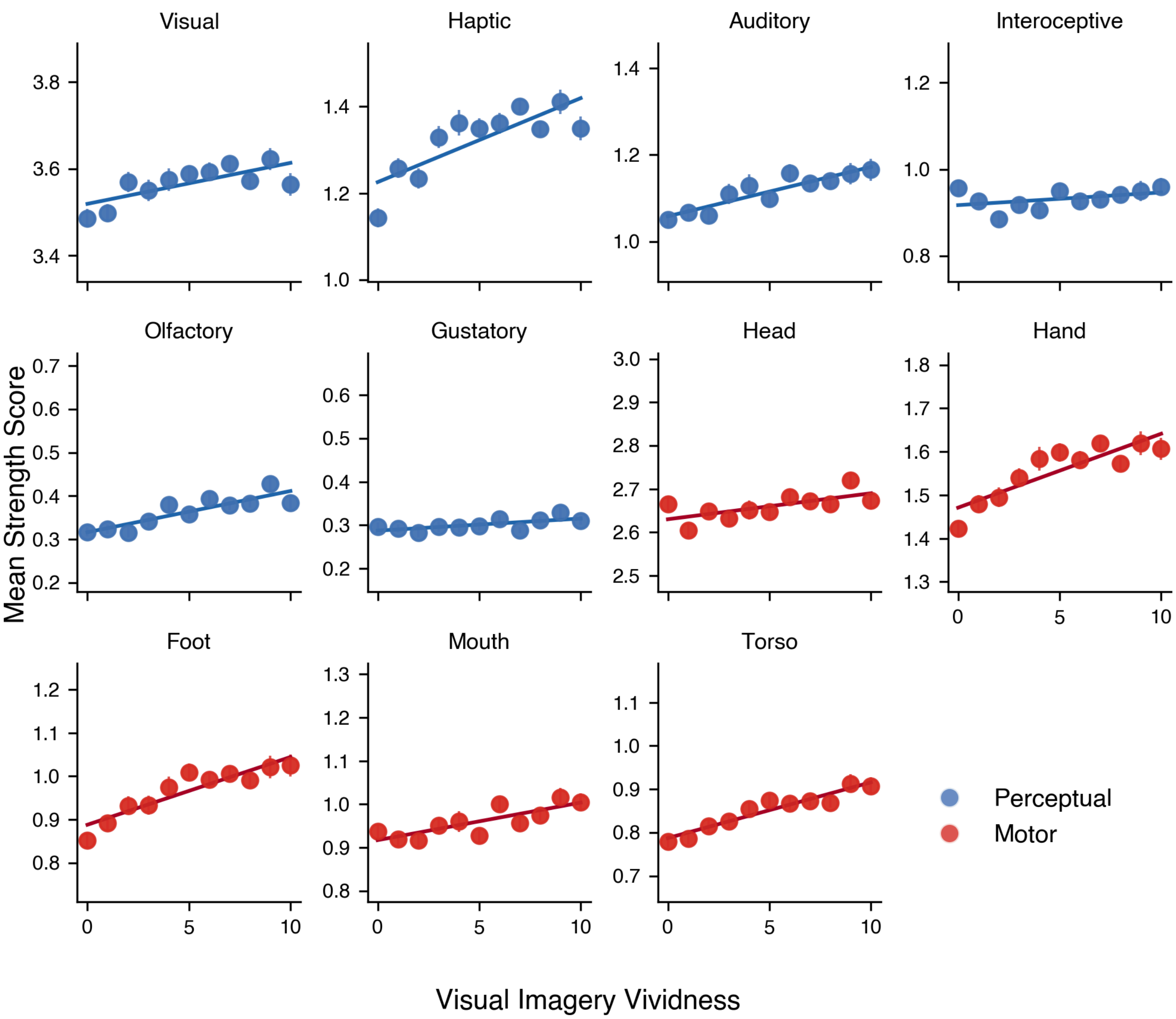

**Figure S5. Individual sensorimotor dimensions by visual imagery vividness**. Individual relationships between visual imagery vividness and each of the 11 Lancaster Sensorimotor dimensions. Top panels show the six perceptual modalities (visual, haptic, auditory, interoceptive, olfactory, gustatory) and bottom panels show the five motor dimensions (head, hand, foot, mouth, torso). Points represent mean scores ± standard error for each vividness level. Regression lines show linear trends. Perceptual dimensions are shown in blue, motor dimensions in orange/red. Note the varying y-axis scales across dimensions, reflecting different baseline frequencies of sensorimotor content in hallucination descriptions.

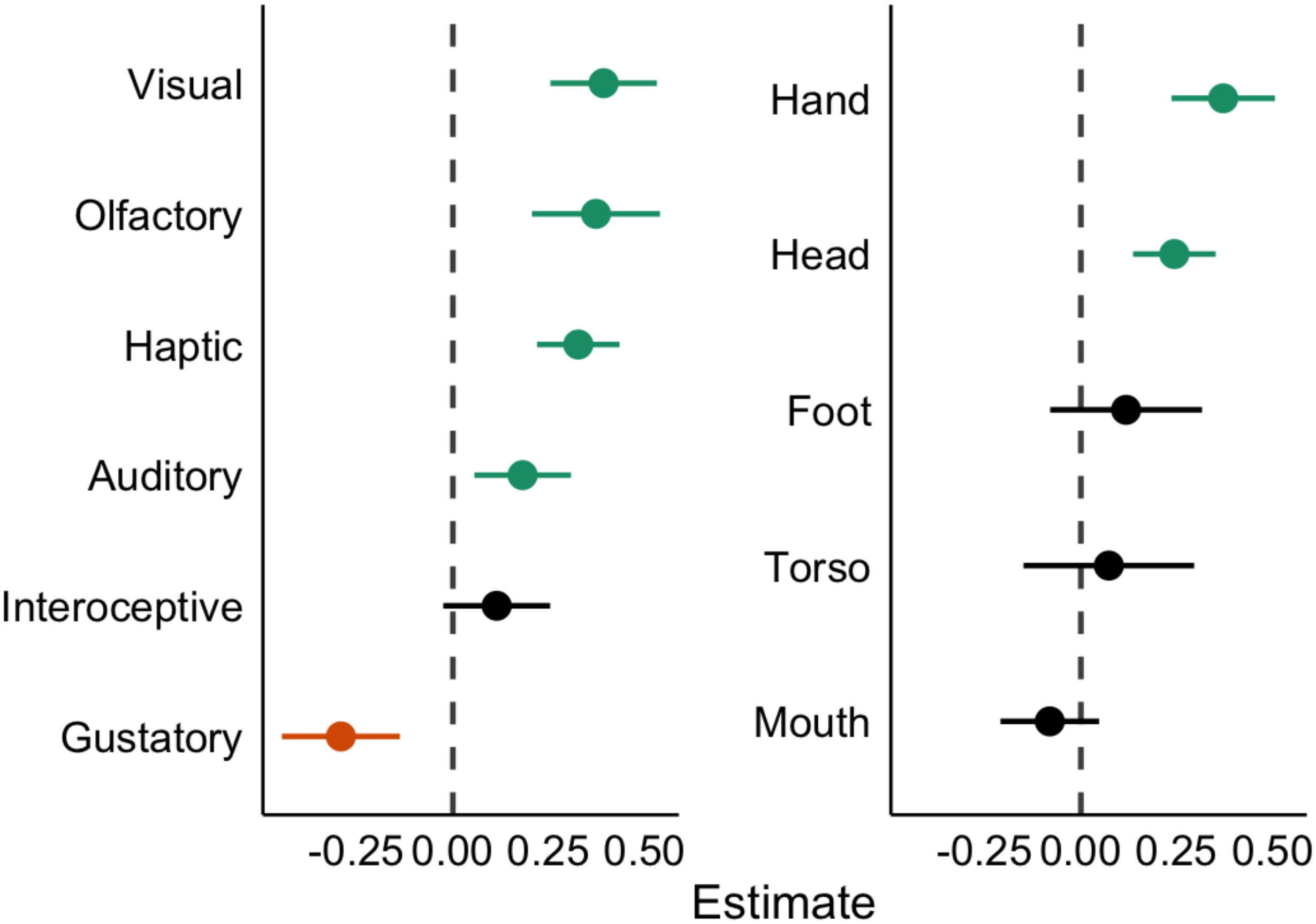

**Figure S6. Sensorimotor effects remain unchanged after controlling for gender.** Effect sizes (β coefficients ± 95% CIs) from perceptual (left) and motor (right) Lancaster models including gender as a covariate. Teal points indicate significant positive predictors of imagery vividness, orange indicates significant negative predictors, and black indicates non-significant predictors. All core sensorimotor effects observed in the primary models remained robust after controlling for gender.

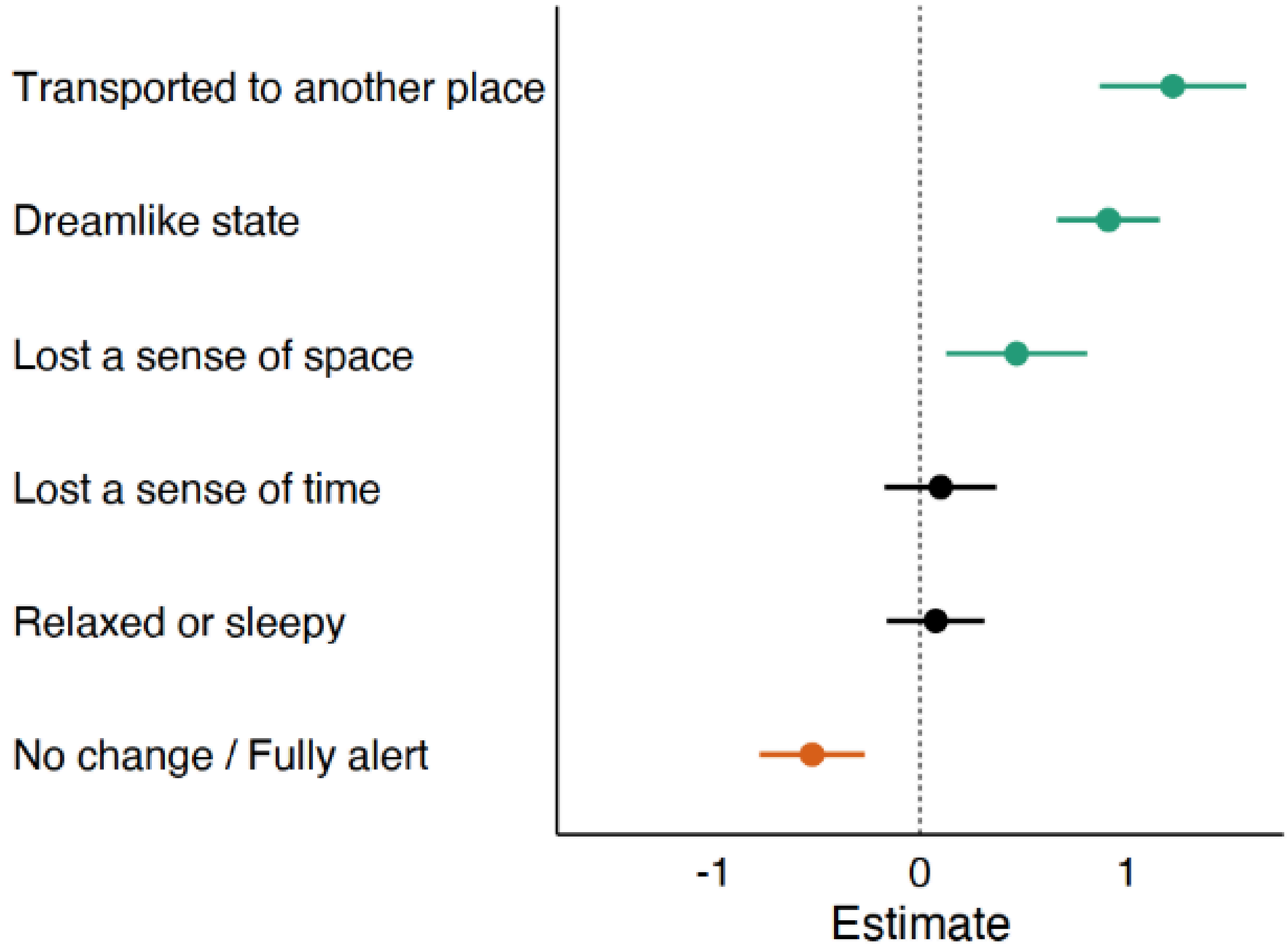

**Figure S7. Altered consciousness states during Ganzflicker are associated with stronger visual imagery.** Coefficient plot showing regression estimates (points) and 95% confidence intervals (horizontal lines) for the association between consciousness change categories and visual imagery vividness. After completing the Ganzflicker paradigm and providing descriptions, participants answered "How did your normal state of consciousness change during the experience?" and could select multiple options from six categories. Green points indicate statistically significant positive associations, and the red point indicates a significant negative association. Participants who reported feeling "Transported to another place" ($p < .001$), experiencing a "Dreamlike state" ($p < .001$), or having "Lost a sense of space" ($p < .01$) had significantly higher imagery vividness scores. Conversely, those reporting "No change / Fully alert" had significantly lower vividness scores ($p < .001$). "Lost a sense of time" and "Relaxed or sleepy" were not significantly associated with vividness scores. This suggests that individuals with more vivid internal visual experiences are more likely to report altered states of consciousness during Ganzflicker, possibly reflecting their more immersive and engaging experience with internally generated visual content.

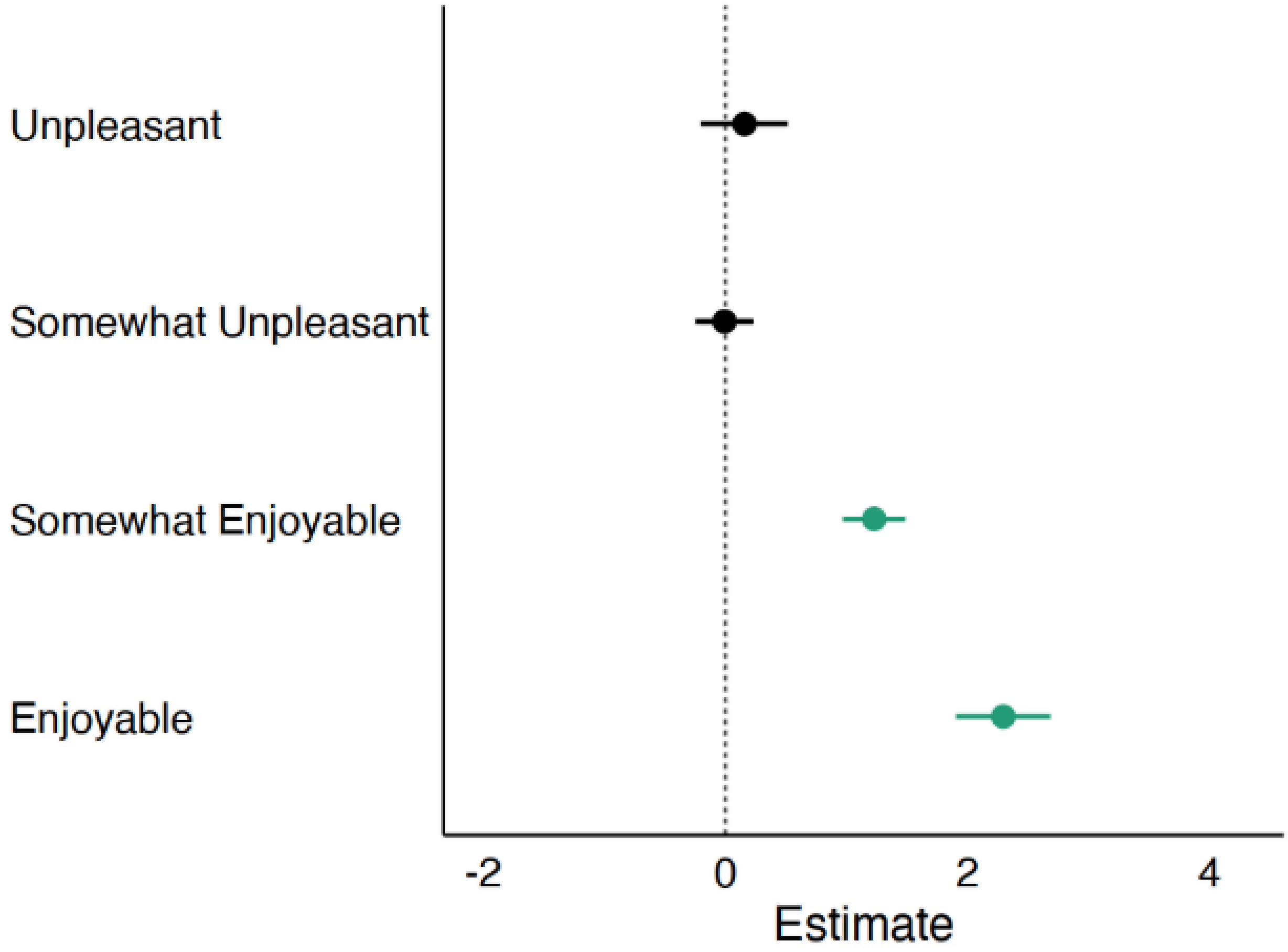

**Figure S8**. **Participants with stronger visual imagery report more positive emotional evaluations of Ganzflicker experience.** Coefficient plot showing regression estimates (points) and 95% confidence intervals (horizontal lines) for the association between emotional evaluation categories and visual imagery vividness, relative to neutral responses ("No feeling about it one way or the other"; reference category). After completing the Ganzflicker paradigm, participants answered "How did you find the experience emotionally?" with response options ranging from "Unpleasant, wanted it to stop" to "Enjoyable, didn't want it to stop." Green points indicate statistically significant positive associations. Participants who rated the experience as "Somewhat enjoyable, but easy to stop" had vividness scores 1.23 points higher ($p < .001$), while those rating it "Enjoyable, didn't want it to stop" had vividness scores 2.29 points higher ($p < .001$) compared to neutral responses. Unpleasant categories showed no significant difference from neutral. This suggests that individuals with more vivid internal visual experiences tend to find visually stimulating paradigms more emotionally engaging and positive.